\documentclass[Afour,sageh,times]{document_style}
\usepackage{moreverb}
\usepackage[bookmarksopen,bookmarksnumbered,citecolor=black,urlcolor=black]{hyperref}

\usepackage{graphicx}
\usepackage{amsmath} 
\usepackage{amssymb} 
\usepackage{amsfonts}
\usepackage{comment}
\usepackage{lipsum}
\usepackage{color}
\usepackage{caption}
\usepackage{subcaption}
\usepackage{algorithm,algorithmic}

\DeclareMathOperator*{\argmin}{argmin}

\newcommand\BibTeX{{\rmfamily B\kern-.05em \textsc{i\kern-.025em b}\kern-.08em
T\kern-.1667em\lower.7ex\hbox{E}\kern-.125emX}}
\begin{document}

\runninghead{Learning Modular Robot Control Policies}
\title{Learning Modular Robot Control Policies}
\author{Julian Whitman\affilnum{1}, Matthew Travers\affilnum{2}, and Howie Choset\affilnum{2} \vspace{-2em}
}
\affiliation{\affilnum{1}Department of Mechanical Engineering, Carnegie Mellon University\\
\affilnum{2}The Robotics Institute, Carnegie Mellon University}
\corrauth{Julian Whitman.
Carnegie Mellon University. 5000 Forbes Ave. Pittsburgh, PA, 15213, USA. Email: jwhitman@cmu.edu
}

\begin{abstract}

Modular robots can be rearranged into a new design, perhaps each day, to handle a wide variety of tasks by forming a customized robot for each new task. However, reconfiguring just the mechanism is not sufficient: each design also requires its own unique control policy. One could craft a policy from scratch for each new design, but such an approach is not scalable, especially given the large number of designs that can be generated from even a small set of modules. Instead, we create a modular policy framework where the policy structure is conditioned on the hardware arrangement, and use just one training process to create a policy that controls a wide variety of designs. Our approach leverages the fact that the kinematics of a modular robot can be represented as a \textit{design graph}, with nodes as modules and edges as connections between them. Given a robot, its design graph is used to create a \textit{policy graph} with the same structure, where each node contains a deep neural network, and modules of the same type share knowledge via shared parameters (e.g., all legs on a hexapod share the same network parameters). We developed a model-based reinforcement learning algorithm, interleaving model learning and trajectory optimization to train the policy. We show the modular policy generalizes to a large number of designs that were not seen during training without any additional learning. Finally, we demonstrate the policy controlling a variety of designs to locomote with both simulated and real robots.

\end{abstract}


\maketitle

\section{Introduction}

A \textit{general-purpose} robot can handle a wide variety of tasks, but such a robot is still fictional.
On the other hand, a \textit{special-purpose} robot can perform a single task, but such a robot is typically limited to that task. 
We believe that modularity can break the trade-off between generality and specialization in robot design, by allowing a designer to combine a small set of building blocks into a specialized robot for a wide variety of tasks. 
In other words, a designer can use modules to create a variety of special-purpose robots. 

\begin{figure}[tb] 
     \centering
      \includegraphics[width=.95\linewidth]{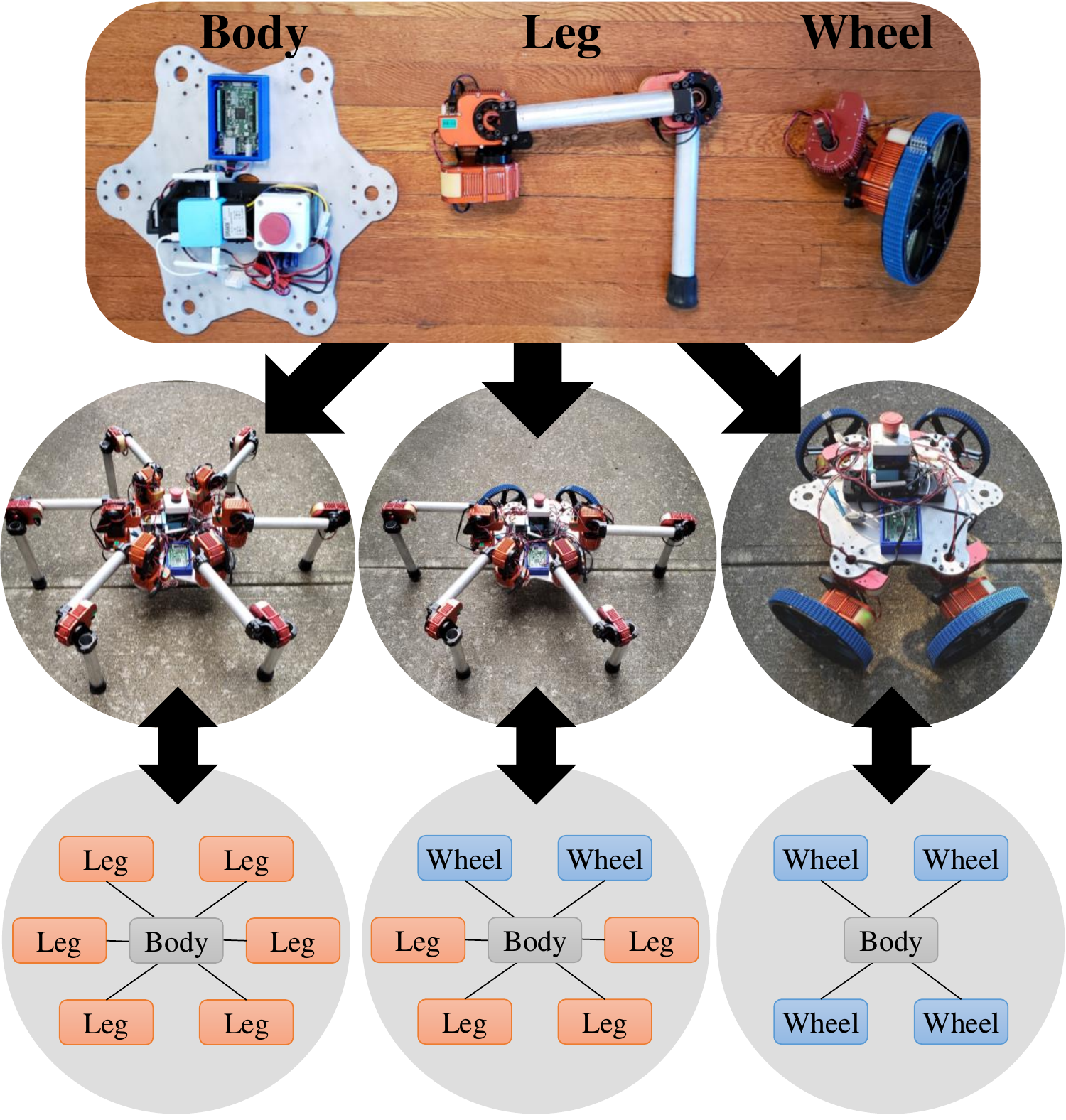}
      \vspace{-0.75em}
      \caption{
      A set of modular components, a body, legs, and wheels, (top) can be combined to form many robot designs (middle). These designs can be represented by graphs (bottom).
       Our modular policy learning algorithm leverages the graph structure common to all such modular designs, enabling us to control any robot composed of these modules.  
     \vspace{-3.5em}
      }
      \label{fig:leg_wheel_body_modules}
\end{figure}

Even with a small set of modules, there is a combinatorial exponential explosion in the number of specialized robot designs that can be generated from that set \citep{chen1994theory, yim1995locomotion, seo2019modular}. Each of these many designs needs a control policy to coordinate motion among its constituent modules.
Further, there is an added nuance to this scaling: how the modules interact with each other also plays a role in optimizing policies.  
Each module needs to behave differently depending on its \textit{context}, i.e. its functional role in the system.
For instance, the legs in a hexapod must behave differently than those same legs need to behave in a quadruped.
And, even within any one robot, the location of the module impacts its desired behavior, e.g. a leg should function differently when used as a front or rear limb.
Dynamic interactions among the different portions of a robot (e.g. coupling between legs) are important within the policy as well. 

Given that context matters, with an eye towards mitigating computational complexity, we aim to create policies for the modules that can automatically determine their function within their given context.
Just as one can install a new keyboard into a computer without reinstalling the operating system, we seek for the module's policies to have a similar ``plug and play'' feature.
However, we seek a ``plug and \emph{adaptive} play'' because in robots, a module's behavior must adapt to its context within the system. 

Instead of optimizing separate policies for each individual robot design, this work develops a learning process that trains policies for the modules.
The modules' policies learn to adapt to different contexts, such that they can be connected together into system-wide policies for various designs.
In our architecture, the global policy (comprised of the union of the module policies) consists of a set of deep neural networks, where there is one network for each type of module, and all modules of the same type share network parameters. 
For instance, a hexapod uses one network six times for its six legs, whereas a leg-wheel hybrid uses that same leg module network for all of its legs and a different network for each of its wheels.
Once trained, the policy can readily transfer to many different robots made from the same set of modules.

Our approach leverages the fact that kinematic structure of any modular design can be represented as a graph, where modules are nodes and electromechanical connections between them are edges. 
Fig.~\ref{fig:leg_wheel_body_modules} depicts the relationship between modular designs and graphical structures which we adopt in this work. 
When a robot design graph is input, the module policies combine to act as a reactive feedback control policy for that robot, where the module policies are connected with a graph structure corresponding to that of the hardware. 
This makes the architecture a \textit{hardware-conditioned} policy, in which the output behavior changes depending on the robot design \citep{chen2018hardware}. 
Introducing a graph structure into a hardware-conditioned policy mediates how robots with different designs share neural network parameters.
We find that this graph-based approach to parameter sharing results in more effective behaviors than non-graph counterparts.

The policy graph structure also enables modules to adjust their behavior to their context. 
Module policies do not act independently, but rather learn to automatically adapt their outputs via a communication procedure \citep{scarselli2008graph} in which they send and receive information over the graph edges.
Some prior work has applied graph structure to robot learning, treating each joint in the robot as a node to learn dynamics \citep{sanchez2018graph} or policies \citep{wang2018nervenet, pathak19assemblies, huang2020smp}.
However, these approaches were shown only in simulation, partly owing to large number of samples needed for training.

To efficiently train the policy to handle a variety of robot designs, we develop a new model-based reinforcement learning (MBRL) algorithm for modular robots.
We choose MBRL over related model-free methods because it has been shown to be more computationally efficient \citep{nagabandi2018neural, chua2018deep, rajeswaran2020game}, which is a requirement for real robots. 
The main innovation of our algorithm is that where prior MBRL methods generate and learn from data for a single robot, we developed a training method that learns from multiple designs simultaneously.
We are able to achieve some notion of scalability, and test on many additional designs, because each type of module, regardless of which robot it is situated in, shares the same dynamics model and policy neural network parameters.
Each module type (e.g. a leg), uses the same parameters in all positions within a single robot, and over all robots that are generated; this prevents them from over-fitting to a single positions and design, forcing them to learn to communicate with other modules in the system and adapt to their context. 

Our algorithm is inspired by the ideas of Guided Policy Search \citep{levine2013guided,levine2014learning,zhang2017deep, chebotar2017path}, which iterates between learning an approximate model of the system dynamics, predicting optimal trajectories under the learned model, and distilling those local trajectories into a global policy.
We achieve policy generalization to new robots: we show zero-shot transfer (i.e. application without additional learning or optimization) to an order of magnitude more designs than were in the training set.
Lastly, we demonstrate our modular policy applied to designs with different combinations of legs and wheels locomoting on real robots.

\section{Related Work}\label{sec:background}
Our modular policy architecture and training algorithm draw inspiration from a range of literature on robot modularity, model-based control, deep reinforcement learning, and methods that learn a single policy for multiple robot designs. 
Because our methods are applied to ground robot locomotion, we focus our attention on related work using such systems.

\subsection{Modular robots}

Modular robots have a rich history in prior work which have posed many interesting questions.
The choice of what constitutes a module is itself an open question.
Modules can have a ``simple'' structure like rotary joints, prismatic joints, links, and brackets~\citep{yim2000polybot, chen2006automatic,kalouche2015modularity, ha2018computational,li2019tri,  liu2020optimizing, grimminger2020open, liu2020motion, SchunkWebsite}.
Alternatively, modules may contain more complex multi-degree-of-freedom structures, like a leg or arm~\citep{chitta2005dynamics,wolfe2012m,kim2017snapbot,gim2020snapbot, liu2021smores}. 
Modules may be all of the same \emph{type,} i.e., homogeneous units~\citep{grandchallenges2007,sproewitz2009roombots,romanishin20153d, daudelin2018integrated,liang2020freebot, liu2020motion, liu2021smores}, or they can be of different types, i.e., heterogeneous units~\citep{stoy2010self, kalouche2015modularity, whitman2017generating, ha2018computational, liu2020optimizing, SchunkWebsite}.  
We therefore take the broad view that modules can be any combination of structure, power, computation, sensing, and actuation, and leave it up to the engineer to define what each module and each module set contains. 
In this work we will define three module types: a leg, steered wheel, and body, from which a variety of locomoting robots can be constructed.

Previous work controlling modular locomoting robots used a library of policies that are either hand-crafted \citep{yim2000polybot,chitta2005dynamics, kalouche2015modularity,daudelin2018integrated} or optimized individually for each robot design \citep{whitman2017generating, ha2018automated, hafner2020towards}. 
While these methods have been successful for individual designs, such approaches do not scale to a large set of designs.
Conventional control policy methods, where highly-trained experts carefully hand-tune the policy over long periods of time for individual robots, become expensive when the robot takes on a new design every day.

\subsection{Model-predictive control}

Recently, promising results in legged and leg-wheel robot locomotion have been made through model-predictive control (MPC) \citep{winkler2018gait, giftthaler2018towards, geilinger2018skaterbots, bjelonic2020rolling, bledt2020extracting}.
These methods use hand-engineered models of the robot dynamics. They repeatedly alternate between optimizing control actions over a finite horizon and executing a short sequence of those actions.  
One drawback of these methods is that they often rely on assumptions specific to an individual robot design; 
for instance, manually setting the sequence of foot or wheel contact modes \textit{a priori}, and assuming centroidal dynamics with massless limbs.
Such assumptions make it difficult to apply these methods to arbitrary modular designs, each of which may have a different number and types of limbs, and a different distribution of mass.

Another drawback of these MPC approaches is that they require highly efficient trajectory optimization subroutines, in order to react quickly to disturbances or unmodelled interactions with the environment \citep{bledt2020extracting}. 
This requirement for efficiency is further complicated by the curse of dimensionality.
To help these methods converge in real-time, pre-computed and engineered heuristics can be used \citep{bledt2020extracting, bledt2020regularized}, but which are customized specifically to an individual design. 
Thus, we turn to learning-based approaches to create reactive controllers, which are comparatively inexpensive to compute at run-time, and where a single algorithm can apply to a variety of designs without assumptions or heuristics specific to the design.

\subsection{Reinforcement learning}

Reinforcement learning (RL) has proven effective at discovering control policies for articulated locomoting robots.
RL can be broadly divided into two classes: Model-free RL (MFRL) and Model-based RL (MBRL).

MFRL has been successful in creating a range of behaviors for many robot morphologies both in simulation and reality \citep{ hwangbo2019learning, xie2020learning, RoboImitationPeng20,  heess2017emergence, simtoreal2018, ha2020learning, hafner2020towards}. 
In MFRL, a control policy is learned directly from an agent's interaction with its environment, collecting and learning from trajectory data (states, actions, and cost/rewards), without explicitly modelling the system dynamics.  
However, these approaches often suffer from high sample complexity and computationally expensive training procedures.
Consequently, to direct a robot to locomote in multiple directions, often a separate policy is trained for each desired robot heading \citep{xie2020learning, RoboImitationPeng20, hafner2020towards}.

In MBRL, a model of the robot's dynamics is learned, then used within trajectory optimization to create policies \citep{rajeswaran2020game}.
MBRL has been shown to be more sample efficient than MFRL \citep{nagabandi2018neural, chua2018deep, rajeswaran2020game}, that is, uses fewer data samples to gain equivalent proficiency. 
This means that MBRL is often more computationally efficient than MFRL, as gathering data is usually the main computational bottleneck in RL control algorithms.

In MBRL, the learned model can be used within MPC to direct the robot to locomote at various headings and speeds \citep{nagabandi2018neural,yang2020data}, but doing so requires running trajectory optimization in real-time.
As an alternative, reactive control using a neural network to compute control actions is comparatively inexpensive at run-time, even for high degree-of-freedom systems. 
Guided Policy Search (GPS) \citep{levine2013guided,levine2014learning,zhang2017deep,chebotar2017path} is a form of MBRL that can produce such a reactive controller. 
GPS iteratively re-fits dynamics models, uses local trajectory optimization off-line to find a series of local trajectories, then merges those trajectories via imitation learning into a global reactive control policy.
A summary of recent MBRL algorithms can be found in \citep{langlois2019benchmarking}.

\subsection{Learning decentralized and multi-design control}

RL can not only be applied separately to individual designs, it can also be used to train one policy for multiple designs.
This is accomplished by applying one set of neural network parameters to a range of designs, and training the network using data gathered from those designs.
One method to do so is to condition the policy network input on the robot design parameters encoded as a vector \citep{chen2018hardware, schaff2019jointly, luck2020data}. 
Then, the policy can be applied directly to, or fine-tuned to transfer to, a design not seen during the initial training period. 
However, this has only been shown previously where the set of designs all share the same topology.

When robots contain repeated structures, but not necessarily the same overall topology, another method to transfer a policy among multiple designs is to train a decentralized policy.
The policy is applied to each repeated part of the robot, e.g. the joints or limbs reused multiple times within the designs. 
By assigning a policy component to each part of the robot, when that part of the robot is removed or an additional part added, the policy can transfer to the altered design. 
For example, \cite{sartoretti2019distributed} trained a decentralized policy for the legs of a hexapod. This demonstrated that sharing policy information among the legs enabled accelerated training compared to a centralized policy for the full robot, and showed the policy could transfer to robot designs with fewer legs. 
However, each leg acted independently, without any internal coordination between limbs, ultimately limiting its capabilities.

In order to allow internal coordination between decentralized policy components, a more complex function than a conventional deep neural networks is needed.
\textit{Graph neural networks} (GNNs) are one such class of functions.
GNNs are a form of neural network that operate over graphs. They can encode graphical structures into neural networks and share learned knowledge among repeated components \citep{wu2020comprehensive}. 
Unlike more conventional neural networks, which have a fixed input and output dimension, GNNs allow a single set of neural network parameters to process inputs and produce outputs with variable dimensions.
For robotics applications, this means that a single GNN can be applied to a range of robots with different numbers of sensors and actuators, as long as those components can be represented as nodes in a graph.
\citet{wang2018nervenet} introduced ``NerveNet,'' which used GNNs as a control policy. 
Each joint on the robot formed a node in the policy graph.
While NerveNet was not necessarily modular, it was shown to generalize to a limited number of simulated designs not seen in training.
Similarly, \cite{pathak19assemblies} and \cite{huang2020smp} used MFRL to train GNNs where each node controlled an individual robot joint, showing that GNNs can transfer to control systems with  different topologies than were seen during training.  
GNNs have also been used to represent an approximation of the forward dynamics model, such that a single model can be applied to multiple designs \citep{battaglia2016interaction, sanchez2018graph}.
But, the GNNs and training procedures used by prior methods encoded only the connectivity between joints into the graph, without the recognition that groups of joints repeated within the robot (e.g. limbs) or rigid components without joints (e.g. bodies or links) can be reused across different designs.
Further, in order to make the model agnostic to the design, prior methods use maximal-coordinate state representations, for instance by including the full world-frame positions and velocities of each link in the robot to their state, which would make transfer to reality difficult in the face of uncertain state estimation.

\section{Problem Overview} \label{sec:problem}

\begin{figure*}[tb] 
     \centering
     \begin{subfigure}{.52\linewidth}
      \includegraphics[width=.95\linewidth]{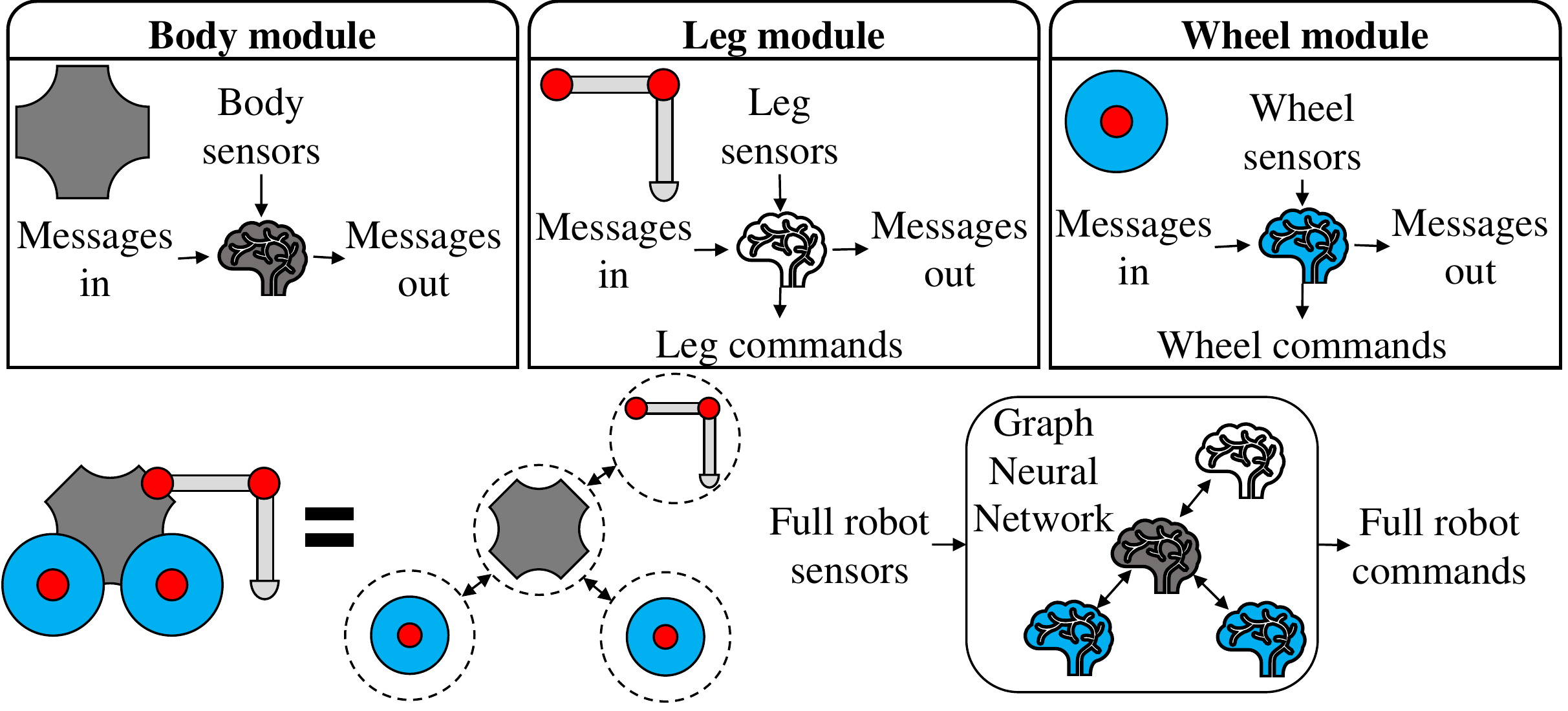}
           \vspace{-1em}
      \caption{}
    \end{subfigure}
\vline\hfill
\begin{subfigure}{.44\linewidth}
     \centering
     \vspace{1em}
      \includegraphics[width=.95\linewidth]{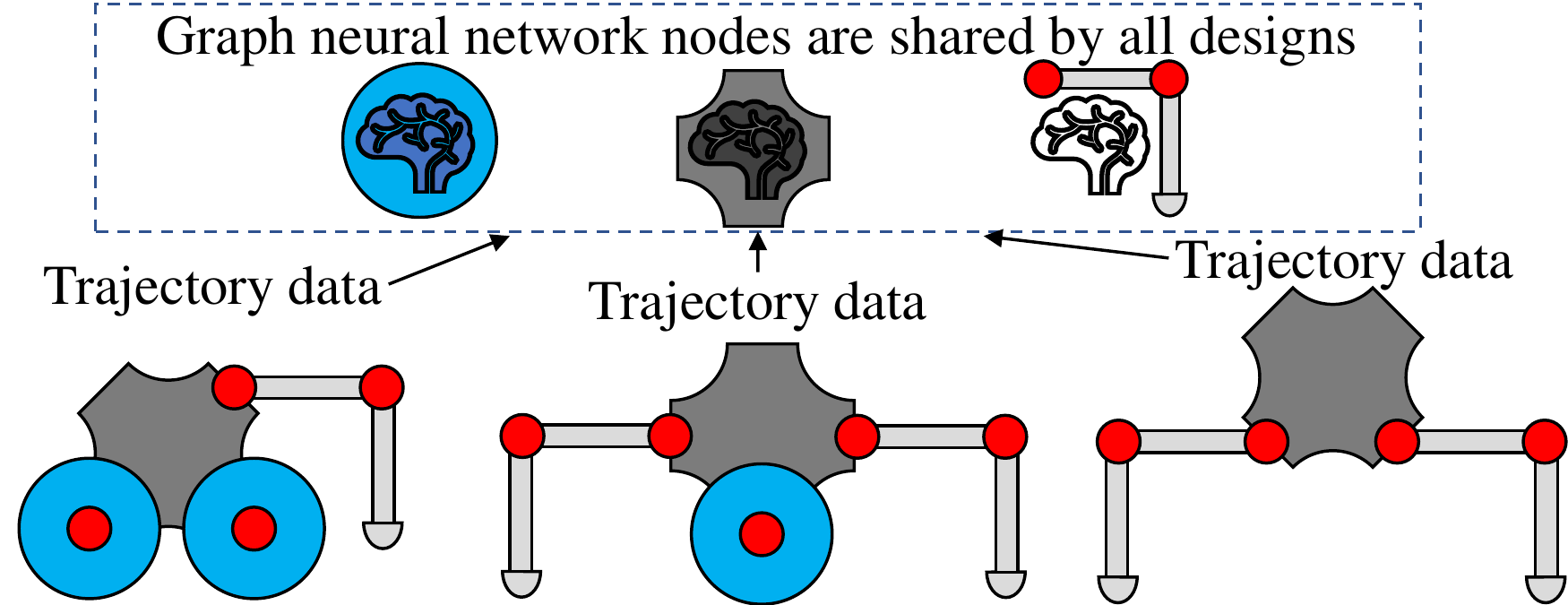}
      \caption{}
      \end{subfigure}
\caption{ 
Our modular policy architecture: (a) the modules, depicted by the three boxes in the upper left, can be composed to form different designs. 
Each module type has a deep neural network associated with it, indicated by the brain icons, which processes that module’s inputs (sensor measurements), outputs (actuator commands), and messages passed to and from its neighbors. 
Assembling those modules into a robot (bottom left) creates a graph neural network (GNN), with a structure reflecting the design, where nodes and edges correspond to the modules and connections between them.
The architecture is decentralized in form, but due to messages passed over the edges that influence the behavior of the nodes, the graph of networks can learn to compute coordinated centralized outputs.
(b) The GNN nodes (top right) are shared by all designs made from these modules. 
The modular architecture is trained using trajectory data collected from a variety of designs (bottom right) such that it can apply to any combination of the modules. 
}
 \label{fig:gnn_cartoon}
\end{figure*} 

In this section, we first define the modules and design graph.
Then, we describe the objective function used to optimize modular policies for locomoting robots.

\subsection{Module and design graph}

We represent each robot system with a design graph, where a node corresponds to a module, and an edge corresponds to the electromechanical connection between two modules.
In this work, the example modules we use to ground the discussion are a two-DoF steered wheel, a three-DoF leg, and a rigid body with no actuation. 
A design $d$ has $N_d$ modules, and can contain multiple modules of the same type.
For example, a robot made up of a body, two wheel, and four leg modules, uses all three module types and has $N_d = 7$ total modules in it. 

Let $M$ represent the number of types of module, where each type has an index $m = \{1, \dots, M \}$; here we will use $M=3$ for the leg, wheel, and body modules. 
Note that this index refers to the type of module, and not an instance of that module in the robot; a single module type may appear multiple times within a design. 
Each module naturally has a state, makes observations through its sensors, and executes actions through its actuators. Let $x_m \in \mathbb{R}^{n_{x,m}}$ be the state of a module type $m$ where $n_{x,m}$ is the size of that module's state vector. Likewise, let $o_m \in \mathbb{R}^{n_{o,m}}$ and $u_m \in \mathbb{R}^{n_{u,m}}$ be the observations and actions, respectively, for module type $m$ where $n_{o,m}$ is the number of observations and $n_{u,m}$ the number of output actions for module type $m$.
Observations contain partial and/or noisy state measurements.
The full states $x \in X$, observations $o \in O$, and actions $u \in U$ of any design are the union of states, observations, and actions of that design's component modules.
Note that the dimensionality of these spaces, for any given design, vary depending on the modules in that design.

Each module type has a $N_{\textrm{ports},m}$ ports, which form connections between modules, and exchange power and data. 
Each port has at most one edge connecting to one neighboring module, or that port may be empty if no module is attached to it. 
In our module set, the body has six ports, and the wheels and legs each have one port.

\subsection{Modular policy optimization problem}
\label{sec:optimization_problem}

During training, our objective is to obtain reactive policies optimized for a set of $K$ modular robot designs $D = \{ d_1, d_2, \dots d_{K} \}$.
The $K$ designs may be fewer than the total number of possible combinations of modules. 
In this work, these designs are assumed to be given, and concurrent work addresses automatic selection of the design training set.  

In order to specify locomotion with various headings and speeds, we input a target velocity to the policy in addition to the robot's sensor observation inputs.
We call this auxiliary input the policy ``goal'' $g \in G$, the target desired velocity for the robot to achieve, represented by a linear velocity ($v_{x}, v_{y}$) and yaw angular velocity ($\omega_z$), so $g = [v_{x}^\text{des}, v_{y}^\text{des}, \omega_{z}^\text{des}], G \subseteq \mathbb{R}^3$.
During training, these desired velocities are sampled from a distribution $g \sim \mathcal{G}$. $\mathcal{G}$ is a distribution over $G$, e.g., a uniform distribution over a bounded range of velocities. 
At deployment time, goals can come from a user (e.g. joystick teleoperation) or from a high-level planner.
Without loss of generality, goals could also contain other desired state conditions.
The policy $\pi: O \times G \rightarrow U$, conditioned on a design, takes an observation and goal as input and outputs actions for all modules in the design. 
We condition the policy on a desired body velocity, which has not been shown by previous MBRL methods.
The policy $\pi$ takes the form of a GNN with parameters $\theta$, which will be described in detail in the next section.

The overall objective of policy optimization is to minimize a cost function $C: X \times  U \times  G, \rightarrow \mathbb{R}^+$.
The cost penalizes deviations of the velocity from the desired velocity, e.g. $C(x,u,g) = || [v_{x}, v_{y}, \omega_{z}] - g ||^2 + c(x,u)$. 
The cost function also includes additional penalties $c$ include regularizing the control input norm, rate of control variation, as well as the roll, pitch, and height of the body. Further cost function details are listed in the Appendix. 
The policy optimization problem can be written as
\begin{equation} \label{eqn:full_optimization}
     \theta^*=  \argmin_{\theta} 
     \underbrace{ \vphantom{\sum_t^T} \mathbb{E}_{g \sim \mathcal{G}} }_{\substack{\text{Expectation} \\ \text{over goals}}} \bigg[ \overbrace{ \frac{1}{K} \sum_{d \in D} \underbrace{\sum_{t=1}^{T} C \big(x_t, \pi_{\theta}(o_t, g), g \big)}_{\text{Cost for individual design } d} }^{\text{Average over } K \text{designs in }D} \bigg].
\end{equation}
Over the course of a trajectory of length $T$, the state evolves according to an underlying forward dynamics transition $x_{t+1} = f(x_t, u_t)$.
The dynamics $f$ are different for each design. 
We assume $f$ is not known analytically, but robot-environment interaction data can be accessed from a simulation.
We develop a GNN architecture to approximate $f$ and represent $\pi$.

\section{Graph neural networks for modular robots}
\label{sec:GNN}

The modular policy and approximate model are implemented as GNNs \citep{scarselli2008graph, wang2018nervenet, wu2020comprehensive}, deep function approximators comprised of a network of neural networks. 
In our implementation, each node in the GNN has a node type corresponding to its associated module type.
For any design $d$, the connectivity of the GNN is set to match the connectivity of the physical hardware graph with nodes $\{ \nu_1, \dots, \nu_{N_d} \}$. 
Fig.~\ref{fig:gnn_cartoon} illustrates the GNN architecture.
The functions mapping inputs to outputs (a.k.a. the neural network ``forward pass'') for a GNN are more complex than they are for conventional multi-layer perceptrons (MLPs, also known as dense neural networks).
MLPs process inputs by sequentially passing vector-structured data through a series of layers.
GNNs must use a more sophisticated series of internal functions to operate on graph-structured data.

\subsection{Graph neural network internal functions}
\begin{figure}[tb] 
     \centering
      \includegraphics[width=.99\linewidth]{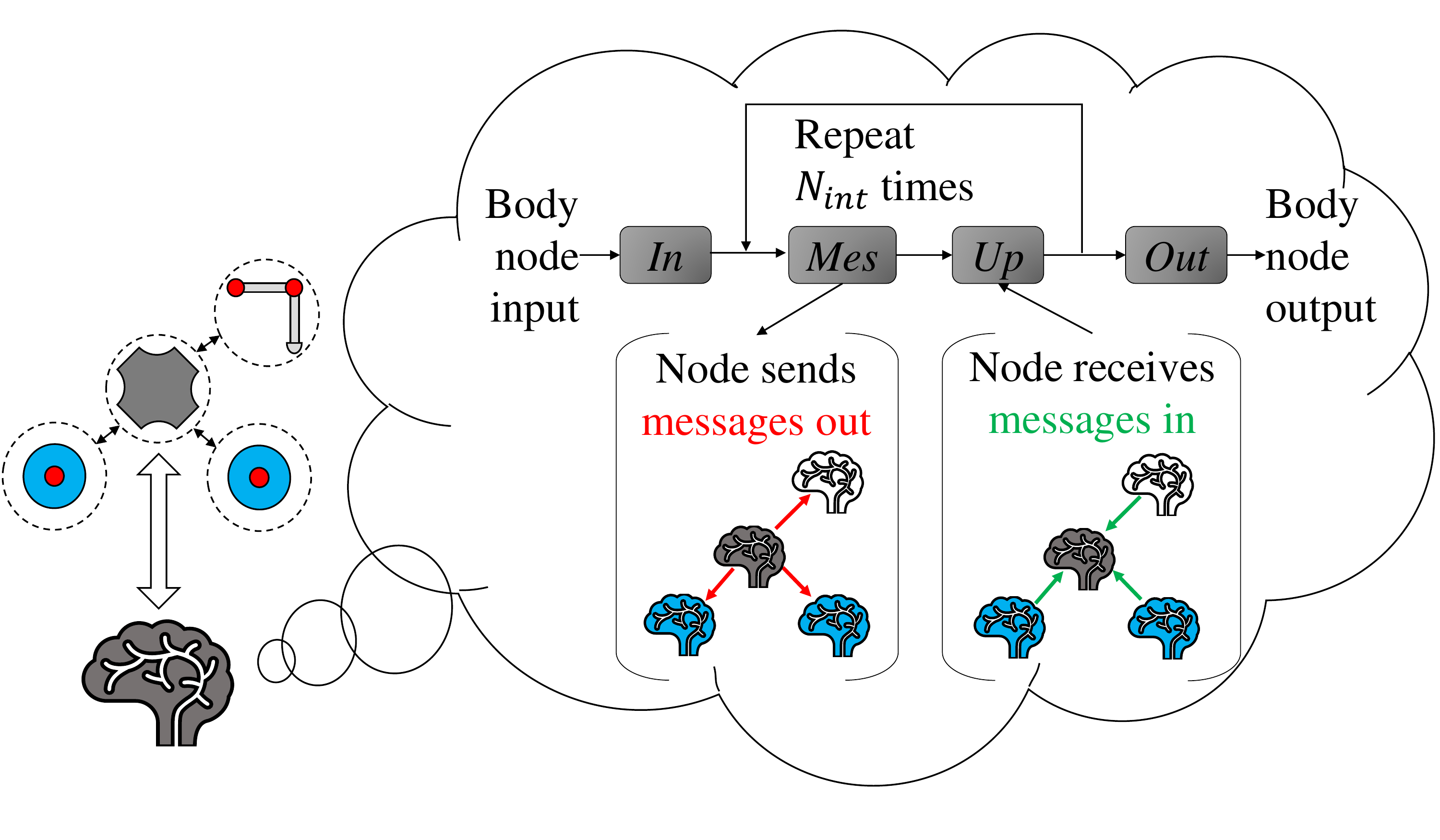}
      \caption{ An illustration of the graph neural network from the point of view of the body module node, depicted as a dark gray brain icon. 
      Each module in the robot (left side) has a graph node, which undergoes a ``forward pass'' indicated by the contents of the thought bubble. 
      The node first obtains the relevant input (e.g. sensor observations from the body). Then, within the space of a single time step in the real world, all nodes compute a series of $N_{\textrm{int}}$ internal propagation steps. During these steps, the nodes exchange messages to propagate information through the graph. 
      All nodes undergo these steps at once, then compute outputs (e.g. control actions for each modules' actuators).  
      See Sec. \ref{sec:GNN} for further descriptions of these functions.
      }
      \label{fig:gnn_internal}
\end{figure}

The GNN forward pass uses a series of functions: first an input function, then multiple internal steps with message-passing and internal update functions, then lastly an output function. The form of our GNN is inspired by \citep{wang2018nervenet}.
Fig. \ref{fig:gnn_internal} illustrates the process of a forward pass from the perspective of the body node, and Algorithm \ref{algo:gnn} describes it in pseudeocode.

\subsubsection{Input function}
At each time step $t$, each node $\nu$ receives an observation $o_\nu$, which is passed through an input function $F_\textrm{in}$ to produce an hidden state vector $h_\nu^0 = F_\textrm{in}(o_\nu)$. Here we use the subscript to indicate that a vector belongs to the node $\nu$. 
Each node maintains its own hidden state $h$. 
Nodes take as input their local parts of the full robot's observation in minimal coordinates. 
For example, a leg module node takes in the local joint angles and velocities from a leg's three joint encoders, but does not require information about the Cartesian position of that limb. 
The body node takes in its orientation, linear, and angular velocities from the IMU sensors on the body.

\subsubsection{Message-passing propagation}

In the GNN, there are two notions of ``time.'' The first is the standard time step discretizing the dynamics and controls, during which the GNN forward pass occurs.
The second notion occurs inside the span of each real-world external time step, when multiple computation steps occur inside the GNN during a single forward pass. 
In the space of one external time step, the GNN computes a series of internal propagation steps to pass messages (real-valued vectors whose content will be learned) between nodes.
This learned communication protocol occurs internally to the network during each time step; it provides a means for the nodes of the GNN to produce collective coordinated outputs.
Each module learns to alter its behavior depending on the messages it receives, and learns to pass messages that inform other modules how to alter their behavior, to achieve the full robot's goals. 

At each internal propagation step, each node sends outgoing messages, receives incoming messages, and then uses those messages to update its hidden state. 
Specifically, after the input function, the graph undergoes $N_{\text{int}}$ internal propagation (message-passing) steps within a single time step.
Let $e$ represent an edge connecting $\nu$ to a neighboring node.
At internal propagation step $i \in \{1 \dots N_{\text{int}} \}$, each node converts its hidden state into outgoing message vectors $\mathbf{m}$ that will be sent over each of its edges using an output function, $\mathbf{m}_{e,\nu} = F_{\textrm{mes},e}(h_\nu^{i})$. 
The superscript on $h$ indicates the internal propagation step index.
The message output function $F_\textrm{mes}$ sends a separate message to each port. 
The content of the messages is a learned output of the node, and not directly human-interpretable.

After each node computes its outgoing messages, each node reads all messages received from its neighbors.
Those messages are concatenated into a vector  $ \mathbf{m}_{\text{in},\nu} = [ \mathbf{m}_{1,\nu} \dots \mathbf{m}_{N_{\textrm{ports}},\nu}]$. 
When a module's port is unoccupied, the node receives zeros as messages over that port.

The maximum number of input and output ports on each node are fixed according to the ports on the modular hardware. 
Then, by concatenating incoming messages, the receiving node can easily learn to determine the source of incoming messages. 
In contrast, recent related work \citep{sanchez2018graph, wang2018nervenet, pathak19assemblies} averaged incoming messages, which prevented the receiving node from determining their source.
Fixing the number and order of messages allows the nodes to implicitly learn to send information about the relative location of the receiver module to the sender, and as a result, allows modules to adapt their behavior according to their placement on the body.

\subsubsection{Update function}
Each node uses the incoming messages to update its hidden state via an update function $h_\nu^{i+1} = F_{\textrm{up}}(\mathbf{m}_{\text{in},\nu}, h_\nu^{i})$. The message computation and internal update functions are called repeatedly for $N_{\text{int}}$ internal propagation steps, iteratively integrating information from incoming messages into the hidden states. 

\subsubsection{Output function}
After $N_{\text{int}}$ internal propagation steps, all of which occur within a single time step, each node computes an output from its hidden state via an output function $F_{\textrm{out}}(h_\nu^{N_{\text{int}}})$. 

\subsection{Implementation}
Each module type has its own instance of the input, update, message, and output functions ($F_\textrm{in}$, $F_\textrm{mes}$, $F_{\textrm{up}}$, $F_{\textrm{out}}$). 
We use MLPs within each of these functions, although other function representations could be used as well. 

An important feature in our application of GNNs to modular robots is that all instances of a module type share the same network weights for the GNN internal functions. 
The policy parameters are divided by module type, $\theta = [\theta_{1} \dots \theta_{M}]$.
Then, $\theta_m$ are the parameters used in functions ($F_\textrm{in}$, $F_\textrm{mes}$, $F_{\textrm{up}}$, $F_{\textrm{out}}$) for all modules of type $m$.
Each module type has the same parameters regardless of the design in which they are used, so the number of learned parameters scales with the number of module types, and not with the number of designs or number of total modules.
When invoked, the GNN nodes are automatically connected to match the design graph, arranging the nodes into the same connectivity as the hardware.  
For example, in a hexapod robot (a body and six leg modules), the GNN contains six leg nodes which all share the same neural network parameters, and a body which has its own parameters.
Each leg module uses the leg-type node parameters to compute their hidden states, messages, and outputs separately.
To properly coordinate full-robot locomotion, the legs learn to alter their behavior according to the messages passed to them via the body module.

\begin{algorithm}[tb]
 \caption{Message passing graph neural network forward pass described by Sec. \ref{sec:GNN}
. Our algorithm uses one GNN as an approximate dynamics model and another as a policy. }
  \label{algo:gnn}
 \begin{algorithmic}[1]
  \STATE Collect graph-structured observation $o$ from robot for the current time step.
  \FOR {$\nu \in  \{ \nu_1, \dots, \nu_{N_d} \}$}
    \STATE Apply input function $h_\nu^0 = F_\textrm{in}(o_\nu)$
  \ENDFOR
  \\ \textit{Message passing internal propagation steps}
  \FOR {$i = 0 \dots N_{\text{int}}$ }
      \FOR {$\nu \in  \{ \nu_1, \dots, \nu_{N_d} \}$}
        \FOR {Each edge $e$ of node $\nu$}
          \STATE Compute message function $\mathbf{m}_{e,\nu} = F_{\textrm{mes},e}(h_\nu^{i})$
        \ENDFOR
        \STATE Send messages to neighbors over graph edges
      \ENDFOR
      \FOR {$\nu \in  \{ \nu_1, \dots, \nu_{N_d} \}$}
        \STATE Aggregate incoming messages, \\ $\mathbf{m}_{\text{in},\nu} = [ \mathbf{m}_{1,\nu} \dots \mathbf{m}_{N_{\textrm{ports}},\nu}]$
        \STATE Apply update function $h_\nu^{i+1} = F_\textrm{up}(\mathbf{m}_{\text{in},\nu}, h_\nu^{i})$  
      \ENDFOR
  \ENDFOR
  \FOR {$\nu \in  \{ \nu_1, \dots, \nu_{N_d} \}$}
    \STATE Apply output function $F_{\textrm{out}}(h_\nu^{N_{\text{int}}})$ to obtain either a next state $x$ (model network) or action $a$ (policy network).
  \ENDFOR
  \STATE Return the graph-structured outputs.
 \end{algorithmic} 
 \end{algorithm}

\begin{figure*}[tb] 
     \centering
      \includegraphics[width=.55\linewidth]{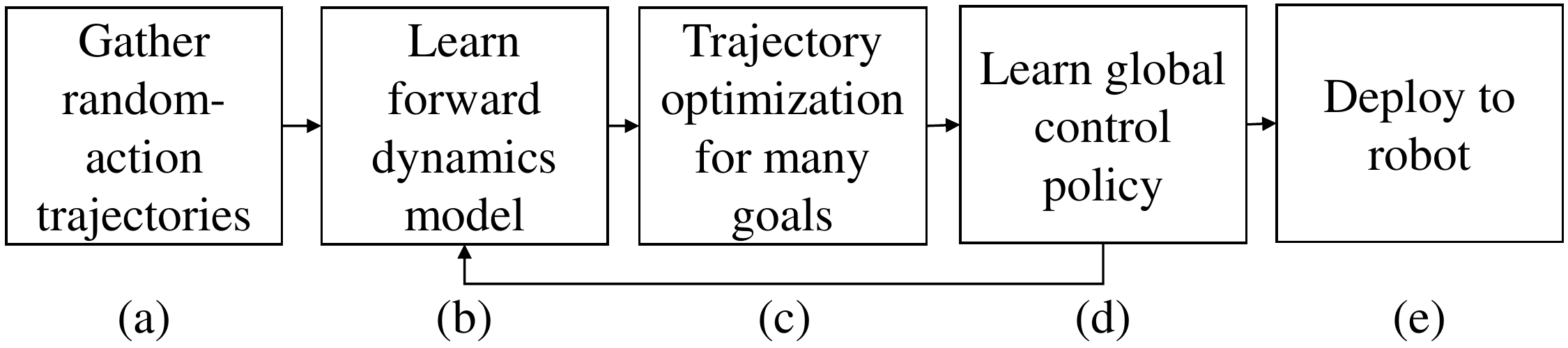}
      \caption{An overview of our model-based reinforcement learning process, described in detail in Sec. \ref{sec:MBRL}. 
      All steps are applied simultaneously to multiple robot designs, which share one set of graph neural network parameters.
      \textit{(a)} First, data is collected for random control actions from which \textit{(b)} initial dynamics approximations are learned.  
       \textit{(c)} Next, the learned model is used to optimize trajectories for the various designs to locomote in a range of headings and speeds. 
       The resulting trajectory data is used to improve the dynamics approximation. 
      \textit{(d)} A global control policy is learned that distills the set of optimized trajectories.
      \textit{(e)} Finally, the policy is tested in simulation, then validated on physical robots. 
      }
      \label{fig:pipeline}
\end{figure*} 

\begin{figure}[tb] 
     \centering
      \includegraphics[width=.95\linewidth]{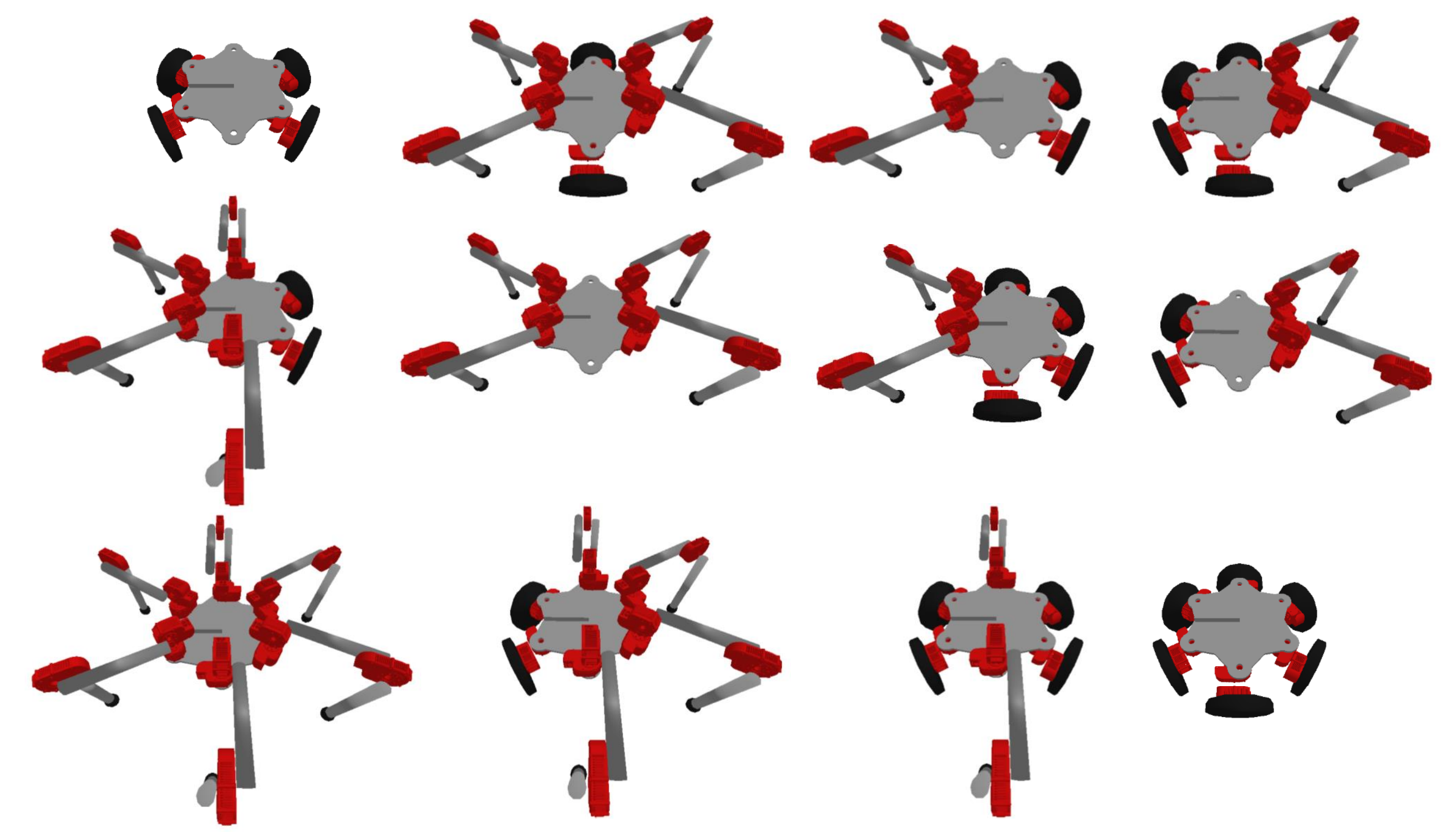}
      \caption{ The twelve ``training set'' designs with different arrangements of legs and wheels used in our experiments. 
      In our learning process, the approximate dynamics models and control policy parameters are shared among these designs, chosen for their bilateral symmetry. 
      The policy can generalize to a much larger ``test set'' of asymmetric designs. 
      }
      \label{fig:twelve_designs}
\end{figure}

\section{Model-based Reinforcement Learning}
\label{sec:MBRL}

We now turn to training the modular policy to produce effective behaviors for a range of designs.
The policy optimization problem (\ref{eqn:full_optimization}) is solved using MBRL.
Our method first learns an approximate model of the system dynamics for use within trajectory optimization (for brevity, henceforth referred to as ``TrajOpt''). 
Then, similarly to Guided Policy Search (GPS), the well-known idea from the MBRL literature \citep{levine2013guided,levine2014learning,zhang2017deep, chebotar2017path}, optimized trajectories are used within imitation learning, resulting in a global reactive control policy. 
Our training algorithm is shown at a high-level graphically in Fig. \ref{fig:pipeline}, and in more detail by Algorithm \ref{algo:mbrl}.

The most significant difference between our algorithm and prior MBRL/GPS methods is that we apply one set of neural network parameters to, and synthesize data from, a varied set of the many possible designs, shown in Fig. \ref{fig:twelve_designs}. 
Two separate GNNs-- a model GNN and a policy GNN-- are trained as part of our MBRL algorithm. 
The model GNN, $\tilde f_{\phi}$ with parameters $\phi$ approximating the forward dynamics, takes the robot state and action as input and outputs the estimated next state.
The model is used in TrajOpt, then refit with data gathered from low-cost regions of the state space visited by the optimized trajectories, becoming more accurate in those regions, such that in the next iteration we obtain trajectories closer to those that would be optimal under the true dynamics \citep{lambert2020objective}.
The policy GNN, $\pi_\theta$ with parameters $\theta$, takes the robot state as input and outputs actions used as control set-points for each module's actuators.

With MBRL, it is possible to learn from real-world robot data  \citep{yang2020data}. 
It is time-consuming and expensive, however, to gather such data from a variety of robot designs. 
We therefore collect all robot data in a simulation environment.
This means training must take additional considerations, described throughout this section, to ensure that the resulting policies can be used on robot hardware.

Our method produces a reactive control policy mapping directly from robot sensor observations to actuator signals.
This stands in contrast to recent MBRL approaches \citep{chua2018deep, yang2020data} that use a learned approximate dynamics model for model predictive control (MPC) to produce actions applied to the robot. 
GPS methods \citep{levine2014learning, chebotar2017path, zhang2017deep} add an additional step, using optimized trajectories to learn a global reactive policy via imitation learning, which is then applied to the robot.
We adopt the latter approach for the following reasons, which overlap with those recently noted by \cite{kaufmann2020deep}:
\begin{itemize}
    \item We use full states within TrajOpt, which includes quantities like body position and velocity, and directly impose a cost on those quantities within the trajectory optimization. 
    But, we learn a global policy that operates over partial observations, mitigating the need to accurately estimate those quantities on a physical robot. 
    \item MPC in real-time can become computationally expensive to run on-board a robot when compared to using a single forward pass of a neural network at each time step. 
    \item The global policy can be used to provide initial seeds to TrajOpt. 
    In our method the approximate dynamics is relearned from the states gathered during TrajOpt.
    This causes the model to become more accurate in regions of the state space near the policy, which leads to better TrajOpt results in the next iteration.
\end{itemize}

In the remainder of this section, we present the process for each step in the algorithm, along with associate experiments demonstrating their efficacy when applied to multiple designs simultaneously.
The hyperparameters for the various components in this method were tuned by hand and can be found in the appendix.

 \begin{algorithm}[tb] 
 \caption{MBRL for modular robots. Each step is conducted for all designs in the training set.}\label{algo:mbrl}
 \begin{algorithmic}[1]
  \STATE Collect dataset $\mathcal{D}$ from random action trajectories.
  \FOR {$i = 1 \dots N$}
  \STATE Learn model $\tilde f_\phi$ from $\mathcal{D}$
  \STATE $\mathcal{D}_{new} \leftarrow \emptyset$
    \FOR {$j = 1 \dots J$}
    \STATE \textit{Trajectory optimization:}
    \IF{$j > 1$}
        \STATE Sample initial state from $\mathcal{D}_{new}$
    \ELSE
        \STATE Use nominal initial state
    \ENDIF
      \FOR {$k = 1 \dots R$}
        \STATE Use current policy $\pi_\theta$ and model $\tilde f_\phi$ to predict the next $T$ actions, $u^0_{1:T}$
        \STATE Use $u^0_{1:T}$ as initial seed for trajectory optimization with dynamics $\tilde f_\phi$ to obtain control $u_{1:T}$
        \STATE Simulate $n_{ex}<T$ steps forward with control $u$
    \ENDFOR
      \STATE Add trajectory from simulation to $\mathcal{D}_{new}$ 
    \ENDFOR
      \STATE  Learn policy $\pi_\theta$ from $\mathcal{D}_{new}$ with behavioral cloning
      \STATE $\mathcal{D} \leftarrow \mathcal{D} \cup \mathcal{D}_{new}$
  \ENDFOR
 \end{algorithmic} 
 \end{algorithm}

\subsection{Initial trajectory collection}
\label{sec:random_traj}

First, trajectories from random actions were gathered \citep{sanchez2018graph, nagabandi2018neural}. 
These trajectories were to learn initial approximation of the dynamics model, and do not resemble the trajectories that are obtained by later stages of the algorithm.
To create smooth random actions for 100 time steps, random actions were chosen for every 10-step interval, and splines were fit to generate actions for every time step in between.
In this work, the actions are target joint velocities tracked by low-level joint PD controllers, but could in general also represent target torques or positions. 
Each design was simulated for a number of trajectories proportional to the number of joints in that design. 
If the robot flipped onto its side (roll or pitch magnitude exceeds $\pi/2$) then that trajectory was ended.
Each trajectory $\mathcal{T} = (x_0, u_0, \dots x_{100})$ was added to a dataset $\mathcal{D}$.

\subsection{Learning modular forward dynamics approximations}
\label{sec:learn_dynamics}

The dynamics model approximation network $\tilde f_{\phi}$ was learned from the trajectories contained in $\mathcal{D}$.
The network learns to approximate the change in state between time steps \citep{nagabandi2018neural, sanchez2018graph, yang2020data},
\begin{equation}
    \tilde x_{t+1} =  x_t + \tilde f_{\phi}(x_t, u_t) 
\end{equation}
where $\tilde x_{t+1}$ approximates the true next state $x_{t+1} = x_t + \Delta x_t$ for a fixed time step $\Delta t$. 
The dynamics of each design is different, but all designs share the same model GNN, trained with batches of data from the training set of designs. 

This model approximation can be learned using standard supervised regression, but additional techniques can increase the accuracy of this approximation in making predictions over multiple time steps.
We adapt two of these techniques to our modular model learning: probabilistic neural networks \citep{chua2018deep} and a multi-step loss \citep{yang2020data}.

\subsubsection{Probabilistic graph neural networks} \label{sec:pgnn}
A probabilistic neural network is one whose output variables are interpreted as the parameters of a probability distribution rather than as a deterministic value \citep{chua2018deep}. 
In our case, this means that the GNN outputs at each node are a mean and a variance of a Gaussian with diagonal covariance, that is,
\begin{equation}
 \tilde f_{\phi} \sim \mathcal{N} \big( \mu_f (x_t, u_t), \Sigma_f (x_t, u_t) \big)       
\end{equation}
Then, the corresponding log-likelihood loss function for a batch of $N$ data points is
\begin{equation}
\label{eqn:single_step_loss}
\begin{split}
    L_f & = - \sum_{n=1}^{N} \log \tilde f_{\phi} (\Delta x_{n} | x_n, u_n) \\ 
    & =  \sum_{n=1}^{N} (\mu_{f}(x_n, u_n) - \Delta x_n)^{\intercal} \Sigma_{f}^{-1} (\mu_{f}(x_n, u_n) - \Delta x_n) \\
    & \quad + \log \det \Sigma_{f}(x_n, u_n).
    \end{split}
\end{equation}
This allows the learned model to capture heteroscedastic noise, and has been found to result in more accurate models even when the data, generated from simulation, is not inherently noisy \citep{chua2018deep, langlois2019benchmarking}.
We found that in practice, it also allows the networks to properly scale the relative loss contributions from state components with different orders of magnitude such that batch normalization as used by \cite{sanchez2018graph, nagabandi2018neural} was no longer necessary.

\subsubsection{Multi-step probabilistic loss}

A learned approximate model is not guaranteed to stay within physically meaningful states when used to predict dynamics over long time horizons \citep{nagabandi2018neural, kolter2019learning}.
To mitigate such divergence effects, one recent approach is to penalize deviations from the ground truth over sequences of states \citep{yang2020data}, rather than from single state transitions as in (\ref{eqn:single_step_loss}).
We adapt the multi-step loss of \cite{yang2020data} for use with a probabilistic network,
\begin{equation}
\label{eqn:ms_loss}
\begin{split}
     L_{f,ms} & =  \sum_{n=1}^{N} 
     \frac{1}{T}\sum_{t=n}^{n+T} \big[ \\
     &     (\mu_{f}(\hat x_t, u_t) - \Delta x_t)^{\intercal} \Sigma_{f}^{-1} (\mu_{f}(\hat x_t, u_t) - \Delta x_t) \\
    & \quad + \log \det \Sigma_{f}(\hat x_t, u_t) \big],
    \end{split}
\end{equation}
where $\hat x$ are recursively predicted states $\hat x_{t+1} = \hat x_t + \mu_f(x_t, u_t)$, and when $t=n$ (the first state in the multi-step sequence) $\hat x_t = x_n$ is sampled from dataset $\mathcal{D}$.
The recursively predicted sequence of states measures the deviation over time of the learned dynamics.
We use sets of $T=10$ sequential states to compute this loss.

One drawback to this approach is that the gradients of this loss become increasingly expensive to compute as the sequence length increases and the network is called recursively multiple times.
This is compounded when using a GNN, which already has more complex gradients than a MLP. 
To reduce training time, we used a form of curriculum learning \citep{bengio2009curriculum}, in which the multi-step sequence length started is incrementally raised over the course of training.
At the start of training, we set $T=1$ in (\ref{eqn:ms_loss}) and then periodically increase it to a final value of $10$. 
This adaptation resulted in the same trained model accuracy with significantly less computation. 

\subsubsection{Training with a set of designs}

$\tilde f_{\phi}$ is trained to approximate the dynamics of the training set of designs.
During training, we sample batches of $N$ state-action sequences (i.e. short trajectories) $\mathcal{T} = (x_n, u_n, x_{n+1}, u_{n+1} \dots x_{n+T})$ for each design, compute (\ref{eqn:ms_loss}), and accumulate the gradients over multiple designs before taking an optimization step with an Adam optimizer \citep{kingma2014adam}.
Averaging the loss over multiple designs prevents the model from over-fitting to any specific design, and instead, to fit jointly to all designs. 

To further reduce computational load, we applied a form of curriculum learning over the number of designs included in each step.
At each training step, a subset of designs were sampled for a forward pass, rather than including all designs in the loss.
The number of sampled designs was incrementally increased until all  training set designs were used at each step.
This adaptation also resulted in the same trained model accuracy with significantly less computation.

\subsection{Trajectory optimization with a learned model}
\label{sec:traj_opt}

The next step in the algorithm is to use the learned model in TrajOpt. 
The goals and states visited by the optimized trajectories will ultimately be used to train a global reactive control policy.
Note that each of these steps is applied at each iteration to all designs in the training set.

\subsubsection{Optimization and Model Predictive Control}
Each TrajOpt solves for a series of control inputs $u$ that minimize an objective function $C$ over a finite horizon length $T$. Each trajectory was given a constant body velocity matching goal $g$ within the objective function, for a minimization problem,  
\begin{equation}\label{eqn:trajopt}
\begin{split}
        u_{0:T}^{*} & = \argmin_{u_{0:T}} \sum_{t=0}^{T} C(x_t, u_t, g) \\
    & \textrm{s.t.} \quad x_{t+1} =  x_t + \mu_{f}(x_t, u_t).
\end{split}
\end{equation}
In the TrajOpt, the dynamics evolve according to the mean predicted by the model GNN.
The initial state $x_0$ in each trajectory has a significant impact on the full MBRL process and on transfer from simulation to reality, as it governs the states used to train the policy, and will be discussed later.

The objective of the TrajOpt process is to create a dataset $\mathcal{D}_{new}$ of ``expert'' demonstrations showing robots tracking various goals $g$ from many initial states $x_0$.
As such, for each trajectory, we sample a new body velocity from the bounded range of goal body velocities $G$.
The cost function $C$ penalizes deviations of the body velocity from the desired body velocity, as described by Sec. \ref{sec:optimization_problem}.
Further penalties in the cost function include costs on the control input norm, as well as the roll, pitch, and height of the body.
A cost term that we found to be critical is the ``slew rate'' penalty, which penalizes abrupt changes in the control inputs. 
While trajectories without this penalty perform well in simulation, they did not transfer well to physical hardware, where actuators perform poorly when commanded to frequently abruptly change direction. 

Many TrajOpt methods exist to find locally optimal solutions to (\ref{eqn:trajopt}).
Prior work \citep{nagabandi2018neural, yang2020data} used simple gradient-free random shooting methods.  
We found that such methods suffer from the curse of dimensionality when applied to high-dimensional systems like our hexapod.
Instead, we turn to a gradient-based method, differential dynamic programming with input constraints  \citep{tassa2014control, amos2018differentiable}, which is able to exploit model linearization to efficiently find locally optimal control inputs. 
Other TrajOpt algorithms could be used as well.
Batches of trajectories were optimized at once using batched forward passes of the learned model.

To create a trajectory, a start state $x_0$ and velocity goal $g$ is sampled, a local solution to (\ref{eqn:trajopt}) is solved under the approximate model.
However, the approximate model is not guaranteed to stay within physically meaningful states when used to predict dynamics over long time horizons \citep{nagabandi2018neural, yang2020data}. 
To mitigate such divergence effects, we combined the multi-step loss described above with TrajOpt in a model-predictive control fashion \citep{nagabandi2018neural, yang2020data}. 
That is, we set the horizon length $T$, solve (\ref{eqn:trajopt}), then execute the first $n_{ex}$ steps of the optimized control in the simulation environment. 
The remaining $T-n_{ex}$ steps are then reused as part of the initial seed for the next replan.
This process was repeated $R$ times for each goal, resulting in each trajectory $\mathcal{T}_{mpc}$ of length $n_{ex}R$. 
The trajectories $\mathcal{T}_{mpc}$ are stored in a dataset $\mathcal{D}_{new}$.

\subsubsection{Initial seeds}

The local TrajOpt requires an initial control input seed, which had a significant impact on the quality of the solution at convergence. 
During the first MBRL iteration, when the global policy is entirely untrained, we used zeros as initial control seeds. 
During subsequent iterations, we use the global policy rolled out on the learned model to create initial control seeds.
This ultimately resulted in lower-cost trajectories than always using zeros as initial control seeds.
More importantly, the optimized trajectories end up nearby the global policy, and the policy is then retrained from those trajectories.
This process iteratively reinforces a consistent gait style.
Without using the policy as the initial control seed, motions generated by TrajOpt were dissimilar between iterations, and cyclical locomotion patterns did not emerge.

\subsubsection{Initial states}

We found the initial state set in the TrajOpt to have a significant impact on the policy's ability to change locomotion heading on-the-fly both in simulation and reality.
While prior GPS learned policies for forward locomotion \citep{zhang2017deep, levine2014learning}, our objective is to learn policies that move the robot in any direction in the plane, and to be usable with tele-operation. 
Consequently, policies must have the ability to quickly change direction and speed.
The policy is learned via imitation, given optimal ``expert'' trajectories demonstrating the robot changing directions.

To create trajectories that contain rapid direction and speed changes, we follow the following steps. 
During the first $M_{init}$ trajectories in each iteration, the initial state $x_0$ was set to a nominal state, standing upright at zero velocity, as shown in Fig. \ref{fig:twelve_designs}.
These trajectories provide expert examples showing the robot starting from rest, stored in $\mathcal{D}_{new}$.
Then, we sample additional initial states from states visited in $\mathcal{D}_{new}$.  
This was inspired by the methods of \cite{zhang2017deep}, who noted that such a process creates overlap in the state distributions visited by the optimized trajectories.
Since the velocity and heading goal of the new trajectory were sampled independently of the goal used to create that sampled state, this resulted in trajectories in which the robots abruptly change directions mid-step, an essential behavior to capture if the end user will be creating the heading on-the-fly with a joystick.
To further enable transfer to reality, we injected a small amount of noise to each sampled initial state, in the form of small perturbations to the joint angles and velocities. 
This provided demonstrations for the global policy of how to optimally recover from disturbances. 

\subsubsection{Gait style objective}

When creating controllers for legged robots, experts often inject their intuition or preferences about gait style.
For instance, recent work on both model-based and model-free reinforcement learning forced cyclical motions (gaits) to emerge by using reparameterization of the actions space \citep{yang2020data} or externally generated cyclical keyframes \citep{xie2020learning, RoboImitationPeng20}.
While our method does not require keyframes for gaits to emerge, we introduce the option to impose a manually selected gait styling with an additional cost on leg joint angles.

The gait style objective was created by first selecting an amplitude, frequency, and phasing for hexapod joint positions that would result in an ``alternating tripod'' step-in-place pattern.
Deviations of any joints in legs from this pattern are then penalized in $C$.
These open-loop joint angles do not move the robot in any direction nor effect the wheels, and the TrajOpt process must discover how to produce locomotion to minimize the velocity matching cost. 
This cost resulted in gaits that follow the main body velocity-matching objective while also remaining near an alternating tripod gait style.
In designs where the gait style may not be the optimal gait pattern (for example, in a quadruped), it can be automatically overcome by TrajOpt because it is set to a smaller weighting than the velocity-matching objective. 
Note that this cost is not necessary for gaits to emerge with our method. 
When a simpler cost on angles deviating from their nominal stance is used, cyclical gaits still emerged which are equally effective in simulation as those learned with the gait style cost.

\subsection{Learning the modular control policy}
\label{sec:behavioral_cloning}

Given a dataset $\mathcal{D}_{new}$ of expert demonstrations, the next step in the MBRL process is to distill these local policies into a global policy via imitation learning.
That is, given many samples of robots moving in many different directions, the imitation learning process acts to ``interpolation'' between the samples.

We use a reactive control policy for the reasons stated at the start of the section: it is simpler to implement on a physical real-time system than is running MPC with the learned model, it is able to operate on partial state observations while allowing the internal trajectory optimization operate on the full state, and it provides initial seeds for TrajOpt.  
We introduce some modifications to the policy inputs and outputs compared to those of related work \citep{zhang2017deep, nagabandi2018neural} in order to facilitate transfer from simulation to reality.

We command target velocities to the actuators, which are tracked by low-level PID loops at a higher frequency on-board the actuator.
The actuators (X-series from Hebi Robotics \citep{HebiRoboticsWebsite}) perform more accurate tracking when provided with a feed-forward (FF) torque value $\tau$.
Thus, in addition to control outputs, we learn an additional output of the policy network that estimates the feed-forward torque needed for the actuator to track the desired velocity.
The data for this output is obtained by tracking the torques experienced by the joints in simulation, contained in $\mathcal{D}_{new}$.

The global policy outputs control command $u$ and FF torque $\tau$, parameterized by means ($\mu_{u}, \mu_{\tau}$) and diagonal variances ($\Sigma_{u}, \Sigma_{\tau}$),
\begin{equation}
  [ \mu_{u}, \Sigma_u, \mu_{\tau}, \Sigma_{\tau} ]= \pi_{\theta}(o, g),
\end{equation}
given an observation $o$ and body velocity goal $g$ as input.
The goal is appended to the body graph node input.

The policy is used deterministically, so at runtime, only $\mu_u$ and $\mu_\tau$ are used.
But, to avoid batch normalization, we found that interpreting the network outputs as a Gaussian (making it a probabilistic GNN, see Sec. \ref{sec:pgnn}) resulted in a more consistent learning process than we found when using a mean-squared error loss.
The policy is learned using a log likelihood  \citep{chua2018deep} behavioral cloning loss, 
\begin{equation}
    \begin{split}
    L_\pi = & \sum_{n=1}^{N} \big[ (\mu_{u,n} - u_n)^{\intercal}\Sigma_{u,n}^{-1} (\mu_{u,n} - u_n) + \\
    & w_\tau (\mu_{\tau,n} - \tau_n)^{\intercal}\Sigma_{\tau,n}^{-1} (\mu_{\tau,n} - \tau_n) \big],
    \end{split}
\end{equation}
for a batch of $N$ samples of $(o, g, u, \tau)$ drawn from $\mathcal{D}_{new}$. $w_\tau$ is a weighting hyperparameter controlling the importance of accuracy of the FF torque predictions relative to the control joint velocity set point outputs.  
A key feature of our method is that the global policy shares data from, and applies to, the full set of modular designs.
That is, the loss over all designs are averaged at each training step.

To further facilitate sim-to-real transfer, we learn the policy with sensor noise and partially-observed inputs.
At each iteration within the policy supervised learning process, we add white noise to the observations.
The body velocity and height, while easily observable in simulation, require state estimation techniques to observe in reality. 
To avoid the added complexity of such state estimation, we remove the body velocity and height from the state observation input. 
To account for latency \citep{yang2020data}, the delay between sensing and actuation, we use the observation from the previous time step as the policy input.
Hyperparameters associated with model learning are listed in the Appendix.

\subsection{Updating the learned model}
The trajectories seen in simulation during TrajOpt form a dataset $\mathcal{D}_{new}$ that provides ``guiding samples'' \citep{levine2013guided} to update the dynamics and learn the policy.
We retrain the model using both $\mathcal{D}_{new}$ and $\mathcal{D}$, adding samples along trajectories relevant to perform effective locomotion without causing catastrophic forgetting of the dynamics in other states.
After the policy is learned, the trajectories in $\mathcal{D}_{new}$ are merged into $\mathcal{D}$, and $\mathcal{D}_{new}$ is reset to empty.
Then, training as described in \ref{sec:learn_dynamics} is continued, warm-started using GNN parameters from the previous iteration.

Note that the TrajOpt problem in (\ref{eqn:trajopt}) uses the approximate model $\tilde f$, so the resulting trajectories are optimal with respect to those dynamics and not to the true dynamics. 
However, the model is relearned from trajectories seen during TrajOpt in the previous iteration, increasing model accuracy in the vicinity of low-cost regions in the state space. 
Subsequently, the optimal trajectories in the next iteration will be closer to the optimum under the true dynamics, and the policy learned from those expert trajectories will be closer to the true solution to (\ref{eqn:full_optimization}).
In other words, iterative process of re-learning the model with data seen during TrajOpt is intended to combat the recently observed ``objective mismatch'' in MBRL: namely, learning a globally accurate dynamics model does not necessarily lead to higher-quality trajectories \citep{lambert2020objective}.

\subsection{Evaluation metric}
\label{sec:eval_metric}

In order to evaluate the quality of the local trajectories and the global policy, we developed a metric quantifying the mismatch between the desired and achieved body velocity over a fixed time period. 
In the case of forward locomotion at maximum speed, as is commonly used in locomotion learning, the distance travelled or average speed serves easily as an evaluation metric. 
In the case of a multi-direction and multi-speed distribution of desired body velocities, (which we represent as a goal distribution $\mathcal{G}$ in (\ref{eqn:full_optimization})), this metric no longer suffices, and we must create a new evaluation metric.  

As an evaluation metric we form a fixed ``test set'' of goal velocities that serve as a finite sampling proxy for the expectation over all possible goals $g \in G$ in (\ref{eqn:full_optimization}).
This test set of goal velocities include moving forward, backwards, left, right, and turning in place left and right.
For each of these test goals, we execute the policy. 
At every $n_{ex}$ step period over the resulting $n_{ex}R$ step trajectory we measured the difference between the desired and achieved average velocity.
Under this metric bounded on $[-1,1]$, higher values are better. A value of 1 would indicate that the robot always moved exactly in the commanded direction, and a metric of 0 indicates that the robot did not move at all.
We did not know in advance how fast each design would be physically capable of moving, and therefore a metric value 1 was not achieved in our experiments, because the desired maximum speed in the goals set was chosen based on a rough estimate of the theoretical maximum robot speed ( additional details and hyperparameters are described in the Appendix).
Similarly, any given design might not achieve the top speed, even were it physically capable of doing so, since the multi-objective TrajOpt cost balances body velocity matching with other costs.
Some designs have difficulty in locomoting in a given direction due to their design, for instance, if a goal velocity requires moving perpendicular to the direction of a wheel.
We use this metric to track the progress of training over the outer MBRL loop and within policy transfer tests.

\subsection{Zero-shot transfer to unseen designs}
 \label{sec:transfer}

The policy and model GNNs were trained on a set of 12 designs, chosen out of the full set of possible module combinations because they are symmetric along their front-back axis and allowed to have the middle port unoccupied.
However, these make up only a small fraction of the total possible space of designs from these modules-- if we allow the designs to be asymmetric, there are an additional 132 possible designs, not seen during training.
The policy can automatically be applied to each of the 132 asymmetric designs. 
We conducted an experiment to test zero-shot transfer (that is, without additional training or modification) of our policy to these designs.

\subsection{Comparison to MLP weight sharing} \label{sec:mlp_compare}
In our GNN architecture, the model and policy are both hardware-conditioned because the structure of the learning representation matches the physical kinematic structure of the robot, and modules of the same type share information.
This stands in contrast to the related hardware-conditioned policies used by \cite{chen2018hardware}, which shared all weights among all robots using an MLP.
To test whether our learning architecture results in higher-quality policies than other weight-sharing architectures using MLPs, we created two baseline comparisons, which we call ``hardware-conditioned MLP'' and ``shared trunk MLP.''

The hardware-conditioned MLP is adapted from the architecture presented by \cite{chen2018hardware}. 
When initializing the network, a fixed maximum number of modules and maximum dimension of the inputs and outputs are specified.
For each robot design the network is applied to, the inputs are padded with zeros to reach this constant maximum dimension length before entering them into the network. 
For example, the leg module has a state dimension size six (position and velocity for each joint), and the wheel dimension three (position and velocity for the first joint, and wheel velocity for the second).
When entering the state of a leg into the network, no zeros need be appended.
When entering the state of a wheel module into the network, three zeros are appended to bring the input size up to the maximum length six.
The output layer of the network is also set to a maximum output size, and the unused outputs for each module are ignored.
The design used during each forward pass is encoded via a one-hot vector, with entries corresponding to the type of each module.
This architecture allows transfer to new designs not seen during training.

The shared trunk MLP is a simpler weight sharing scheme.
The bulk of the neural network weights are shared by re-use of the hidden layers across the 12 training set designs.
To account for the different dimensions of states, actions, and observations among the designs, each design is given its own input and output layers that are not shared. 
Then at each forward pass, the design index is passed to the network, so that the corresponding input and output layers are used.
It is not possible to transfer this network to new unseen designs without further training, as each design has its own input and output layer specific to its dimensionality.

For each of these two network architectures, a separate network instance was used as a model and policy.
The number of layers and depth were tuned to approximately match the capacity and depth of the set of GNN nodes. 
We then applied our MBRL algorithm to measure the policy efficacy via the velocity matching metric.
The gait style objective was used in these experiments.

\subsection{Application to real robots}
\label{sec:real_robots}

Our MBRL algorithm was designed to create a control policy that allows a user to drive a variety of robot designs with a joystick in reality.  
The body contained a battery pack, Ethernet switch, WiFi router, and and IMU. 
A USB gamepad was connected to an off-board laptop, on which joystick inputs were converted to goals, appended to the control policy input, and joint-level commands were computed and sent via WiFi to the robot.

\section{Results}
\label{sec:results}
Our method enabled us to control a variety of robot designs with a single set of GNN parameters.
The modules behave differently when placed in different locations on the robot, even though the GNN node weights are the same for all modules of the same type, because the messages passed between nodes differ depending on the relative position of the modules. 
Further, the modules behave differently within different designs-- for example, the gait pattern that emerges to control the quadrupedal design is different than that of the hexapod.
We ran the algorithm for 3 iterations (alternating between batches of model learning, TrajOpt, and policy learning), which resulted in a policy trained for 12 designs, after approximately 10 hours on a large desktop with 18 Intel i9 cores and four NVIDIA RTX 5000 graphics cards. 
We believe with further code optimization time to train could be reduced, for example, by parallelization of functions that were conducted independently and sequentially for the 12 designs. 

The following subsections describe the results of our experiments on model and policy learning with multiple designs, on zero-shot transfer, and demonstrate sim-to-real transfer.

\subsection{Modular model learning} \label{sec:multidesign_model}

We conducted an experiment to validate the utility of training a shared model with data from multiple three-dimensional articulated robot designs. 
We created datasets using the random-action procedure from Sec. \ref{sec:random_traj} for three designs (as shown in the left-most column of Fig. \ref{fig:twelve_designs}): four-wheel car, a six-leg walking hexapod, and a design with two wheels and four legs.
We then divided the data into training and validation sets, and trained a GNN and a MLP for each design separately using two data regimes: either 100 trajectories or 1,000 trajectories.
The total number of parameters in the GNN and MLP were made comparable in this experiment.
We also trained a single shared GNN model using data from all three designs.

\begin{figure}[tb] 
     \centering
      \includegraphics[width=.99\linewidth]{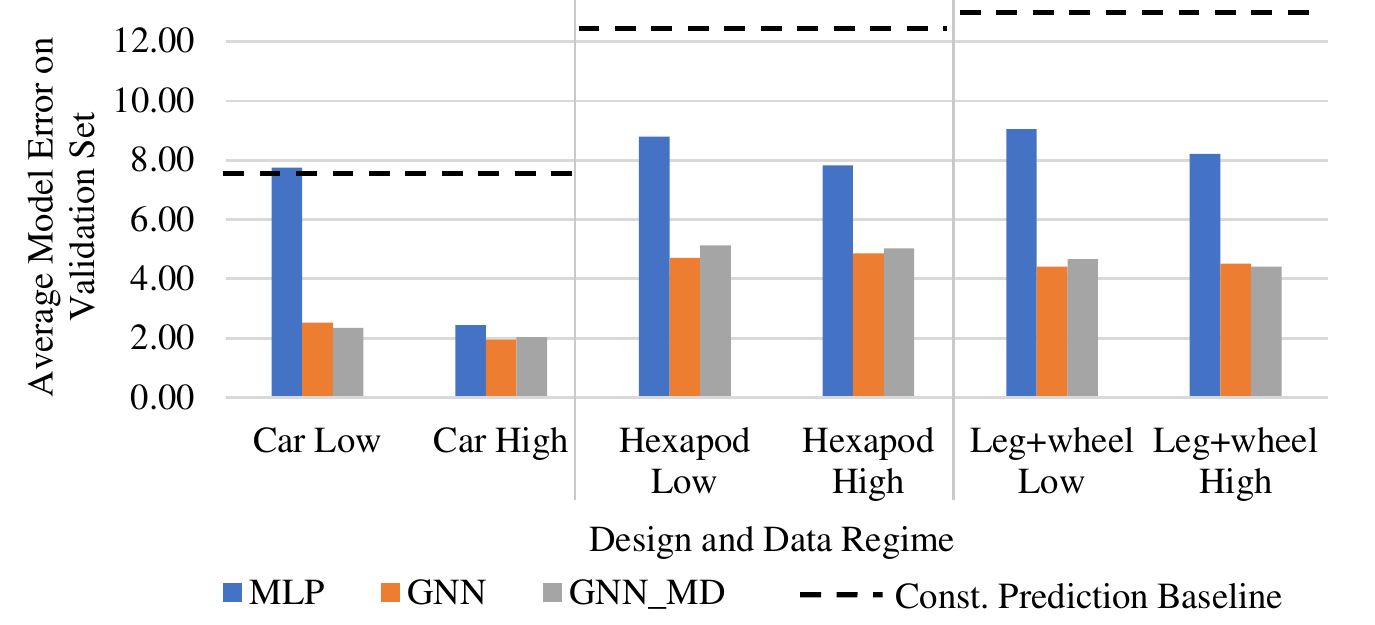}
      \caption{Results described by Sec. \ref{sec:multidesign_model}, where a dynamics model approximation is learned from random state-action data. The graph neural network (GNN) has a lower validation error than a multi-layer perceptron (MLP) with similar capacity and depth. It also is more data-efficient, achieving comparable validation error with $10$ times less data.
      We tested model learning on three designs: a four-wheel car-like design, a hexapod, and a leg-wheel hybrid. Images depicting these three designs are in the left-most column of Fig.~\ref{fig:twelve_designs}. 
      Learning a model from the data of multiple designs (GNN MD) does not harm the validation error. 
      Even though the designs differ in their dynamics due to having different number and type of limbs, the training process also has access to more total data.
      The constant prediction baseline (dashed lines) indicates the the validation error for predictions of zero state change between steps.
}
      \label{fig:model_learning_test}
\end{figure}

We compared the validation error of a forward dynamics approximation network learned via comparably sized GNNs and MLPs within high-data and low-data regimes.
We compared further against a GNN trained using data from all three designs.
We also compared the prediction validation error between the conditions to a ``constant prediction baseline'' \citep{sanchez2018graph}: the error value that would be obtained if the state change prediction was zero at each step. 
The results of this experiment are shown in Fig. \ref{fig:model_learning_test}. 
We found that shared weights between modules of the same type helps prevent over-fitting in low-data regime, and also results in lower validation error for the same number of parameters. Data from multiple modules of the same type contribute to the parameters of the corresponding graph node.
Where \cite{sanchez2018graph} conducted a similar experiment, they used higher-capacity models (e.g. deeper networks) and thus was able to obtain lower validation error than we obtained.
However, lower validation error has recently been shown to not necessarily correspond with the cost of trajectories obtained in model-based TrajOpt \citep{lambert2020objective}.
We also observed that using data from multiple designs had little impact on the validation error, indicating that parameters can be shared among designs to accurately predict the future states of multiple designs made from the same set of components.

\subsection{Modular policy learning}
\label{sec:multidesign_policy}

Next we explored whether a policy created via our full MBRL process with data shared among 12 designs would perform as well as a policy trained with only one design at a time, with the same hyperparameters.
In this experiment, we first use the full MBRL process on 12 designs at once, at each step using data from all designs. 
Then, we use the full MBRL process on the lowest and highest degree-of-freedom systems (car and hexapod) independently, at each stage using only data from a single design.
Table \ref{table:multidesign_policy} shows the evaluation metric applied to the car and hexapod designs; this indicates that the policy trained with shared data between multiple designs is able to perform similarly to a policy trained on only one design.
\begin{table}[ht]

\centering
\begin{tabular}{|c | c | c |} 
  \hline
  Velocity matching metric & Hexapod & Car  \\
  \hline
Trained alone & 0.63 & 0.80 \\ 
  \hline
 Trained with 12 designs
 & 0.73 & 0.80 \\
   \hline
\end{tabular}
\caption{Modular policy training result for Sec. \ref{sec:multidesign_policy}.} 
\label{table:multidesign_policy}
\end{table}

\subsection{Generalization to unseen designs}
\label{sec:transfer_results}

\begin{figure}[tb] 
     \centering 
      \includegraphics[trim={0.1in 0in 0.1in 0in},clip,width=.95\linewidth]{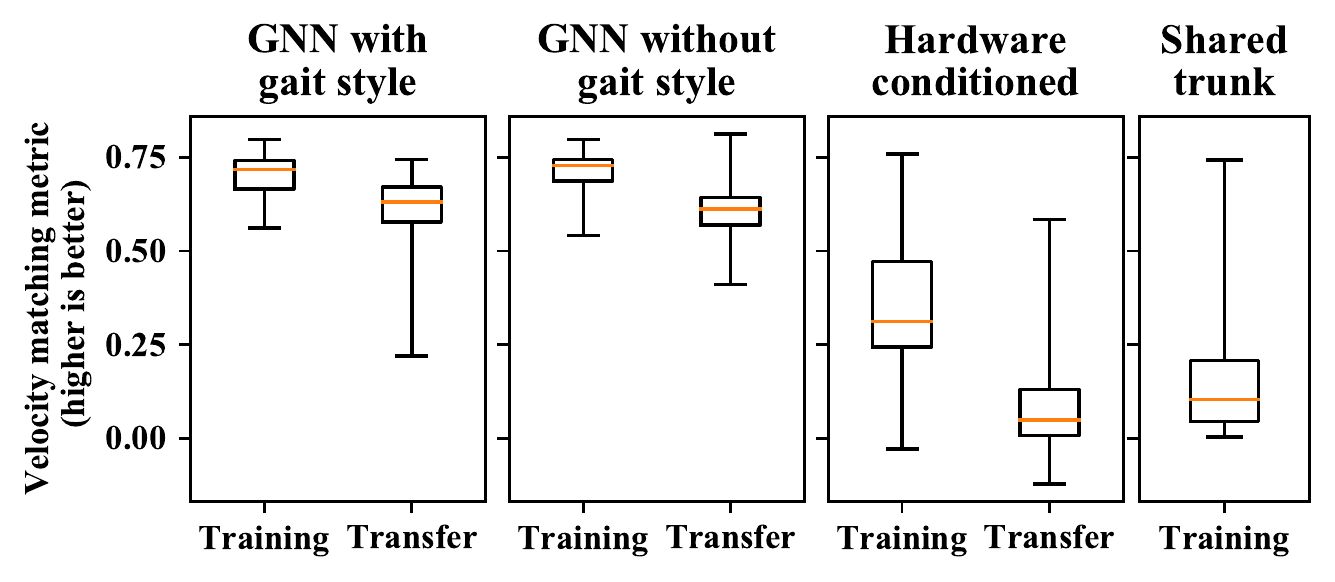}
      \caption{Results of applying the policy to the training and zero-shot transfer set of designs.
      The left and left-center plots show our modular GNN architecture with and without the gait style objective applied, described by Sec. \ref{sec:transfer_results}. 
      The right-center and right plots show the baseline comparisons with multi-layer perceptrons, described by Sec. \ref{sec:mlp_compare}. 
      The training set contains 12 designs, and the transfer set 132 designs not seen during training. The mean of the set is shown in orange, the boxes show the first and third quartiles, and the maximum and minimum of the set are shown by the top and bottom whiskers for each set.
      The policies were measured using a velocity matching metric, where higher values indicate that the policy tracked the desired robot velocity well.
      We found that our modular policy results in effective locomotion for different headings and speeds on a range of different robots, and is able to generalize (without additional training) to an even larger set of designs. 
      }
      \label{fig:transfer_boxplot}
\end{figure}

We trained the control policy with data shared between 12 bilaterally symmetric designs, then tested the policy on 132 simulated asymmetric designs without further training or optimization.
Fig. \ref{fig:transfer_boxplot} shows the evaluation metric, as described in Sec. \ref{sec:eval_metric}, applied to the test set (robots seen during training) and transfer set (not seen during training).
While the average metric for the transfer designs is lower than the training set, we found that the policy was able to direct all designs in the commanded direction on average.
The designs which performed worst in the transfer tests qualitatively appear to be those with fewer limbs and more asymmetries in limb placement, such that their dynamics differed most from that of the designs in the training set. 
The results are similar with and without the alternating-tripod gait style objective applied during training.
Simulated transfer tests on designs not seen during training are included in the supplementary video: \url{https://youtu.be/LTe4LZHpajY}. 

\subsection{Comparison to MLP weight sharing} \label{sec:mlp_results}
The results of our weight-sharing baseline comparisons with MLPs are shown in Fig. \ref{fig:transfer_boxplot}.
We found that the GNN policy had both a higher mean performance and narrower range of performance among both the training and transfer set of designs. 
Although sharing neural network parameters centrally for all parts of multiple robots was demonstrated previously for fixed-base manipulators \citep{chen2018hardware}, we found a hardware-conditioned MLP to be less effective than a GNN for our locomoting robots.
This experiment shows that the inductive bias we applied in learning, that is, matching the structure of the graph network to the structure of the robot, enables more significantly more successful policy learning outcomes than using a more generic architecture.

\subsection{Physical robot validation}
\label{sec:physical_robot_results}

\begin{figure*}[tb] 
     \centering
     \begin{subfigure}{.49\linewidth}
     \includegraphics[trim={0in 0in 0in 0.02in},clip,width=\linewidth]{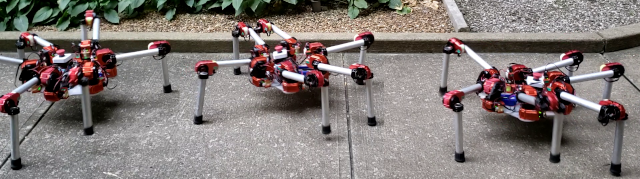}
      \includegraphics[trim={0in 0in 0in 0in},clip,width=\linewidth]{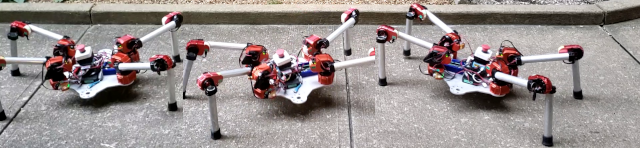}
    \includegraphics[trim={0in 0.02in 0in 0in},clip,width=\linewidth]{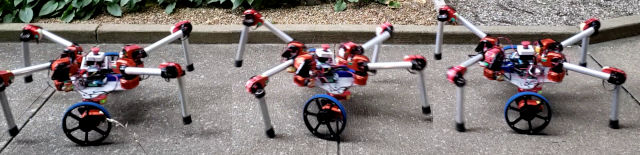}
      \includegraphics[trim={0in 0.05in 0in 0in},clip,width=\linewidth]{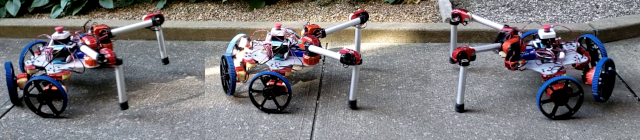}
        \includegraphics[trim={0in 0in 0in 0.025in},clip,width=\linewidth]{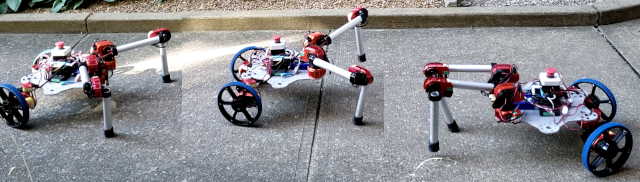}
        \includegraphics[trim={0in 0.03in 0in 0in},clip,width=\linewidth]{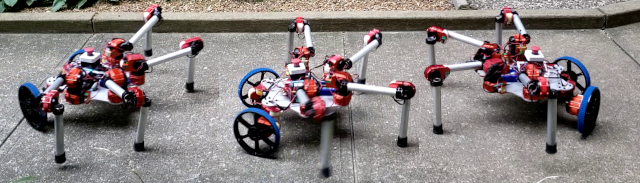}
     \end{subfigure}
        \begin{subfigure}{.49\linewidth}
        \includegraphics[trim={0in 0in 0in 0in},clip,width=\linewidth]{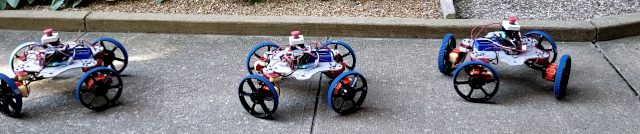}
        \includegraphics[trim={0in 0in 0in 0in},clip,width=\linewidth]{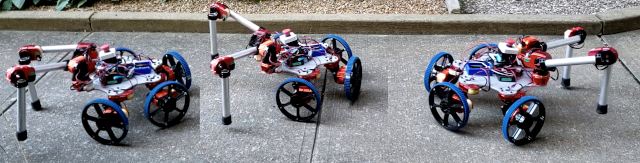}
        \includegraphics[trim={0in 0in 0in 0in},clip,width=\linewidth]{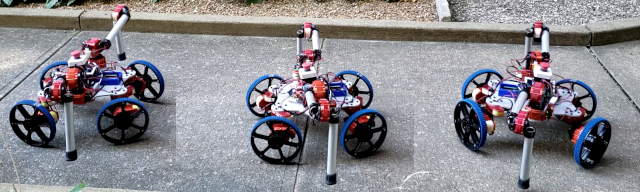}
        \includegraphics[trim={0in 0in 0in 0in},clip,width=\linewidth]{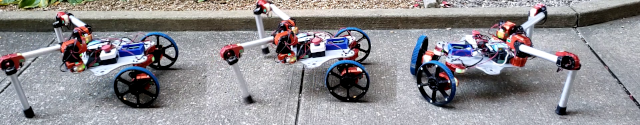}
        \includegraphics[trim={0in 0in 0in 0in},clip,width=\linewidth]{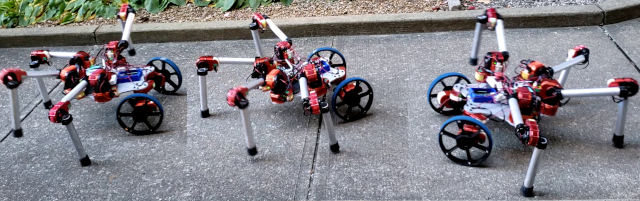}
        \includegraphics[trim={0in 0in 0in 0in},clip,width=\linewidth]{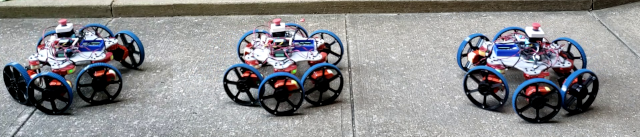}
          \end{subfigure}
      \caption{Timelapses of the twelve modular designs controlled with the learned policy. The robots were teleoperated with a joystick demonstrating forward and turn-in-place locomotion on a sidewalk. Video of the robots can be viewed at \url{https://youtu.be/LTe4LZHpajY}}
\label{fig:timelapses}
\end{figure*}

We control the twelve ``training set'' robots using the modular policy outdoors on a sidewalk. 
The goal heading and speed used as the policy input was commanded via a joystick. 
Fig.~\ref{fig:timelapses} shows a time-lapse of this demonstration, and the videos can be viewed at \url{https://youtu.be/LTe4LZHpajY}.
Qualitatively, most designs performed well, although differences in ground interactions between simulation and reality appear to hamper some of the designs where slipping or dragging contacts occurs.

\section{Discussion}
\label{sec:discussion}
\begin{figure}[tb] 
     \centering
      \includegraphics[trim={0in 0in 0in 0in},clip,width=.9\linewidth]{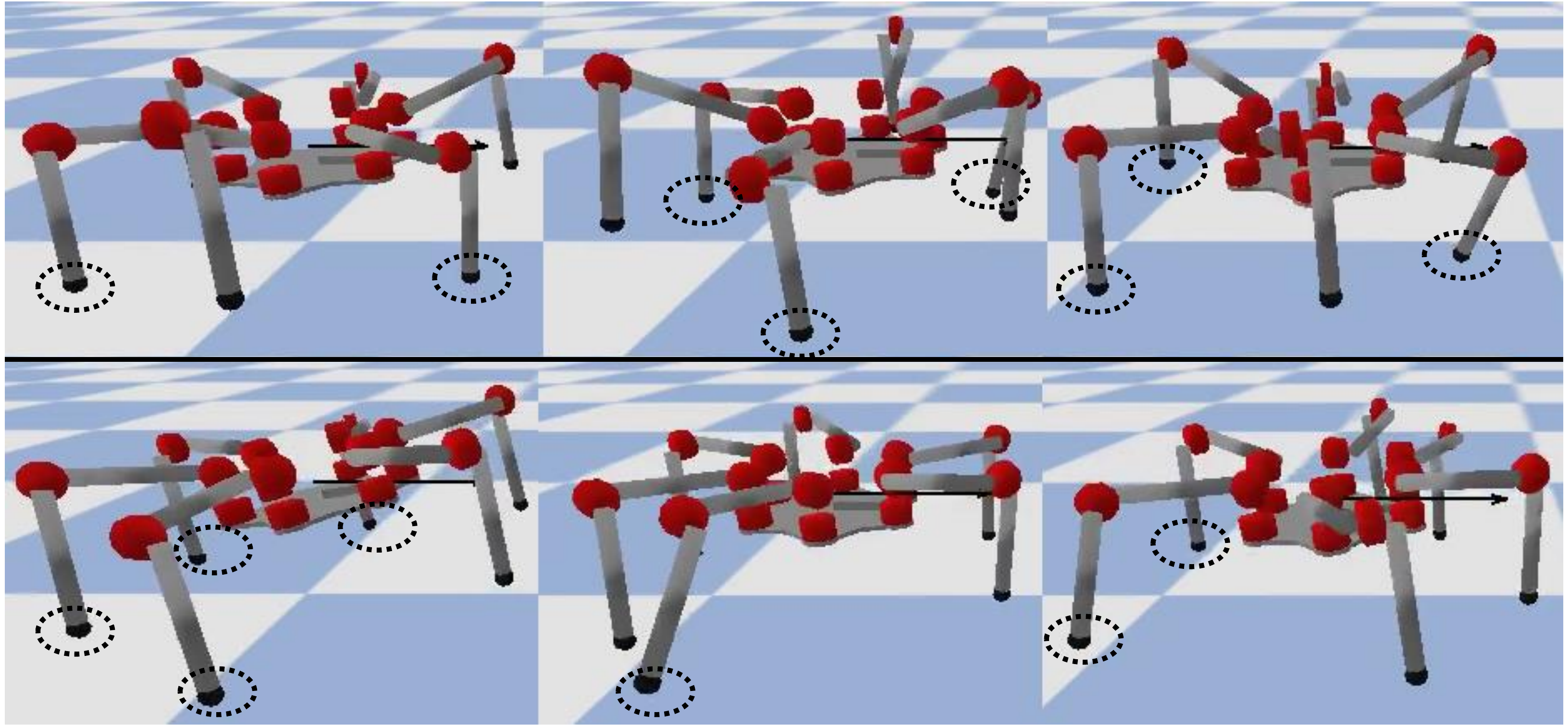}
      \caption{A time lapse of the simulated hexapod motion with (top) and without (bottom) the alternating-tripod gait style objective, walking from left to right. The feet in contact with the ground are circled within dotted lines. Both gaits move at a similar speed, but without the gait style objective, a different contact sequence emerges.}
      \label{fig:gait_style}
\end{figure} 
Model-based trajectory optimization for legged or leg-wheel hybrid robots, such as \citep{winkler2018gait,geilinger2018skaterbots,bjelonic2020rolling,bledt2020extracting}, typically make the assumption that the contact sequence is known a priori.
An emergent feature of our work is that the contact sequence can be discovered automatically.
This becomes particularly important when we are tasked with creating trajectories for multiple robot designs, because each design may have different combinations of legs and wheels, resulting in a different optimal contact sequence.
We added a gait style objective for the case where the user biases learning towards a particular contact sequence.
An example of the difference in contact sequences that emerge with and without the gait style objective is shown in Fig. \ref{fig:gait_style}.
When the gait style impedes learning, it is overcome; we found this was the case with the quadruped, which qualitatively has a similar behavior, and quantitatively has a similar performance measure, both with and without the gait style.

The contact sequences that emerge differ among the designs, even though they are the result of a single policy.
We found that the interplay between trajectory optimization and global policy imitation learning played a key part in enabling the policy to be effective on many designs.
Training the modular policy using the dataset created by each iteration of trajectory optimization creates module-level behaviors that apply increasingly well to the full range of designs. 
In the first iteration of the pipeline, before the policy has been trained at all, the initial control seed in TrajOpt is zero.
The local trajectories that arise for the different designs appear dissimilar, and some are low-quality local optima.
The policy learns to imitate the collective dataset of trajectories, so in the first iteration, the policy may not be effective on even a single design.
However, that policy provides an intial seed for the next iteration of TrajOpt, resulting in lower-cost local minima that are more similar across designs.
We observe that this effect compounds until ultimately the policy is effective on the full training set of designs. 

A number of choices in the algorithm are taken to aid computational efficiency.
We allow for intra-limb coordination of multiple joints encapsulating multiple actuated joints within a graph node, which reduces the number of message passing steps compared to related work \citep{sanchez2018graph,wang2018nervenet,huang2020smp} in which each node controls, or approximates the dynamics of, one joint.
We use smaller capacity models than \cite{sanchez2018graph} to learn the dynamics, but find that these are still effective when used within trajectory optimization, especially when combined with the multi-step loss function introduced by \cite{yang2020data}.
We also showed that the additional inductive bias introduced by shared weights between limbs of the same type helps prevent over-fitting in the low-data regime, resulting in sample efficient training.
With these choices, in addition to some strategic curriculum learning as described throughout Sec. \ref{sec:MBRL}, we were able to conduct training on a single computer without cloud compute resources. 
We believe this to be an important feature in making deep learning accessible and reproducible.

\section{Conclusion}
\label{sec:conclusions}

This paper introduced a model-based reinforcement learning method to control a variety of modular robot designs with a single policy.
Both a dynamics approximation and a global control policy are learned with graph neural networks (GNN) that share parameters among distinct designs and learn from a combination of data from those designs. 
Our GNN formulation embodies a novel inductive bias \citep{battaglia2018relational} in the learning representation and training process relative to prior works \citep{wang2018nervenet, sanchez2018graph,huang2020smp}: not only is a robot made up of a tree of joints, but there are multiple types of modules repeated in the graph structure of modular designs, and the structural modules without joints impact the dynamics and control as well. 
The GNN learns how each module type (body, wheel, or leg) should behave within the context of the other modules present in the design.
As a result, we observe emergent behavior wherein limbs with the same neural network weights behave differently for different designs and locations on the body. 
The policy allows a user to drive a range of robots with a joystick, or for the policy to be used as motion primitives within a high-level path planner.

We showed that our policy transfers readily to designs, composed of those same modules that were not seen during training, without additional learning or optimization. 
We were inspired by the computer vision research community, which has found that the right learning representation (convolutional neural networks) and a diverse set of training examples (dataset of images) enables generalization to images not seen in training \citep{russakovsky2015imagenet}. 
Similarly, we find that for modular robots, a learning representation that stores knowledge about dynamics and controls in the module graph nodes (a GNN), and a diverse set of training examples (designs with various combinations of modules) enables policy generalization to designs not seen in training.

In the development of our methods, we noticed a number of limitations.
As noted by \cite{lambert2020objective}, there is a fundamental mismatch in the functions being optimized in model-based reinforcement learning-- a more accurate model does not necessarily result in a better policy. 
Our use of Guided Policy Search techniques \citep{levine2013guided} appear to mitigate this problem, at each iteration increasing the accuracy of the model in regions of low trajectory cost.
Future work will thus consider convergence analysis, as well as further study on the effect of the number and size (number of joints) of the modules, as well as the effect of the many other hyperparameters on convergence, in particular when applied simultaneously to many robot designs at once.

Another limitation of our work lies in the simulation to reality transfer.
We showed that the policy transfers to reality, but the performance of robots in simulation appears better than in reality.
In future work, we will investigate learning from data collected on a combination of simulation and physical hardware.
Our experiment, with results in Sec. \ref{sec:multidesign_model}, found that a GNN can learn from smaller datasets more efficiently than a comparably sized MLP when data from modules of the same type are shared, which shows that our method has the potential to learn from physical robot data. 
We are also investigating combining our methods with existing sim-to-real techniques like simulated latency \citep{yang2020data} or domain identification \citep{RoboImitationPeng20}.

In this work, we learn a model from simulation, and perform trajectory optimization using that learned model.  
One reason we do so is the potential to learn a model using data from reality either in place of, or in addition to, simulation data. 
A reasonable alternative would be to use the simulation directly for trajectory optimization. 
Our initial attempts to do so using Pybullet \citep{coumans2019} failed, which we attribute to the difficulty in using finite differences to compute dynamics linearizations while frequently making and breaking contacts.
Further, we found it computationally less expensive to compute batches of trajectories in parallel with neural networks on GPUs than with parallel physics simulations.
But, this may be possible using a differentiable physics simulator \citep{carpentier2019pinocchio}, or using gradient-free TrajOpt \citep{williams2018information} in tandem with physics simulations that run in parallel on GPUs \citep{liang2018gpu}.
To account for sim-to-real transfer, the resulting policies could be used to gather real-world data from a robot, and that data used to learn an error correction term to create a model that is hybrid of a simulator and a neural network \citep{ajay2019combining}, potentially rivaling our current sample efficiency by learning only terms defining how reality differs from the simulation.

We note that there are a number of different formulations used for GNNs in recent literature. The GNNs of \cite{sanchez2018graph, wang2018nervenet, huang2020smp} all differ, and \cite{wu2020comprehensive} reviews a variety of additional formulations yet to be applied to robotics. 
The type of network used impacts the forms of data and robot topologies that can be included. 
Future work will investigate the effect of different GNN formulations on model and policy learning.

We presented generalization of the policy to designs not seen during training; however, we noticed that the worst-case from those designs, while still moving in the right direction on average, performed worse than the lowest-performing design from the training set under our evaluation metric. 
One way to address this would be to include some bilaterally asymmetric designs in the training set, such that the policy learns to coordinate limbs in asymmetric designs. 
However, such an approach would likely not be scalable in the general case, as even our small set of components can be used to form over 100 designs. 
In future work, we plan to scale the method up to larger design spaces by sampling designs at each training iteration, rather than using every design in a fixed set at each iteration.
Further, we recognize that not all designs have the physical capability to move effectively, and so we intend to interleave design optimization with policy training to simultaneously identify high-performing designs and create their policies.

This work, though using a limited set of modules and operating in uniform environments, represents a stepping stone towards the vision of rapidly deployable task-specific robots.
So far, our method (and most other MRBL methods of which we are aware) learn a model and operate the robot on flat ground.
Model-free methods have been able to address variable terrain heights \citep{heess2017emergence}.
Future work will experiment with model-free fine-tuning \citep{nagabandi2018neural} or adding additional inputs to the learned model such that the policy can adapt motions based on the sensed environment.

As modular robot hardware components become more commonplace, we believe the need for scalable methods to simulate, prototype, and evaluate the potential of the many possible designs will grow as well.
Our methods and planned future work aim to make modules, both in physical hardware and in their control, into the general building blocks with which a user can specialize the robot to the task.

\section*{Funding}
This work was supported by NASA Space Technology Research Fellowship NNX16AM81H.

\appendix
\section{Hyperparameters and cost functions}

All neural networks in this work were implemented using PyTorch \citep{PyTorch}.
The many hyperparameters listed below were tuned by hand. 
Tuning was conducted on each process independently (i.e. tuning the model learning first on its own, then tuning the trajectory optimization on its own, etc.). 

\subsection{Simulator parameters}
The simulator used throughout this work had the following settings:
\begin{itemize}
    \item Simulator: Pybullet \citep{coumans2019}
    \item Simulator time step: 1/240 seconds
    \item Time steps per control action: 20, resulting in an effective time step for learned model and controller of 20/240 seconds.
\end{itemize}

\subsection{Approximate dynamics model learning}
The dynamics model learning process used the following settings:
\begin{itemize}
    \item Length of random rollouts: 100 steps
    \item Number of random rollouts per design: 300 rollouts per actuated joint on the designs (between 12 and 18 joints).
    \item Batch size per design forward pass: 500
    \item GNN (internal state, message, hidden layer) size: (100, 50, 350)
    \item GNN (input, message, update, output) function hidden layers: (0, 1, 0,0)
    \item GNN Update function LSTM hidden size: 50
    \item Activation function: ReLU
    \item GNN training steps initial: 10000
    \item GNN training steps after each set of TrajOpt complete: 1000
    \item Learning rate: $1 \times 10^{-3}$, decays by half every 2500 steps
    \item GNN weight decay (weight norm penalty):  $10^{-4}$ 
    \item Number of designs used per step: 6, increasing by one every  2000 steps
\end{itemize}

\subsubsection{Translation and yaw invariance}

The dynamics of motion under a constant gravitational field are invariant to the translation and yaw of the system.
Prior work \citep{sanchez2018graph, nagabandi2018neural}  learned the model in the world frame, then subtracted out the body translation to compute network inputs. 
We extended this translation-invariance with an additional inductive bias by recognizing the symmetry of the dynamics with respect to not only the translation in the plane but also the yaw of the body. 

The model dynamics were learned in a yaw-aligned frame centered at the body location.
This frame is different from the conventionally-defined body frame, as the height, roll, and pitch of the body are still relevant when the dynamics occur under the external force from gravity.
The x-position, y-position, and yaw $\gamma$ in the plane were removed from the state.
Then the world velocities were rotated by the negative yaw; for example, the body world-frame linear velocity $v \in \mathbb{R}^3$ was rotated to $v_B = R_z(-\gamma) v$, where $R_z(\cdot)$ represents an SO(3) rotation matrix about the z-axis.
The state transitions (and also the global policy, discussed later) were learned with respect to the yaw-aligned frame. The change in x-position, y-position and yaw with respect to that frame was predicted by $f_\phi$.

\subsection{Trajectory optimization}
The trajectory optimization used the following cost weights:
\begin{itemize}
    \item Horizon length $T = 20$ 
    \item Execute the first $n_{ex} = 10$ steps of each optimized control sequence
    \item Cost on norm control signal: 0.01
    \item Slew rate cost: 7
    \item Cost $w_z||z - z_d ||^2$ for z-position (height) of body: $w_z = 5$,  $z_d = 0.23$
    \item Cost $w_r||r||^2$ for roll of body: $w_r = 30$
    \item Cost $w_p||p||^2$ for pitch of body: $w_p = 20$
    \item Cost $w_x||x - x_d ||^2 + w_y||y - y_d ||^2$ for time-varying x- and y-position of body, following desired goal position over time: $w_x = w_y = 110$,  $x_d = x_0 + v_{x,d}t$  $y_d = y_0 + v_{y,d}t$, given body position $(x_0,y_0) $ at the start of the MPC replanning step
    \item Cost $w_\gamma||\gamma - \gamma_d ||^2 $ for time-varying yaw of body, following desired goal yaw over time: $w_\gamma = 25$,  $\gamma_d = \gamma_0 + \omega_{\gamma,d}t$, given body yaw $\gamma_0 $ at the start of the MPC replanning step
    \item Penalty for joint angle of open-loop gait style for first joint on wheel modules (set to zero joint angle for all time): 0.4
    \item Penalty for joint angle of open-loop gait style on legs (set to zero joint angle for all time for first joint, sine wave with amplitude 0.6 rad and period 1.25 s for second and third joint): 6
    \item Maximum goal velocity in (x,y) directions, based on maximum wheel rotation rate: 0.7 m/s
    \item Maximum goal velocity in yaw directions, based on maximum wheel rotation rate and body radius: 2.4 rad/s
    
\end{itemize}

\subsection{Control policy learning}
\label{sec:policy_appendix}
The control policy learning process used the following settings:
\begin{itemize}
    \item Length of expert TrajOpt rollouts: 40 steps
    \item Number of expert rollouts per design: 750, or 1000 at final iteration
    \item Batch size per design forward pass: 500
    \item GNN (internal state, message, hidden layer) size: (100, 50, 250)
    \item GNN [input, message, update, output] function hidden layers: (0, 1, 2, 0)
    \item GNN Update function LSTM hidden size: 50
    \item Activation function: ReLU
    \item GNN training steps: 8000
    \item Learning rate: $3 \times 10^{-3}$, decays by half every 2000 steps
    \item GNN weight decay (weight norm penalty):  $10^{-4}$ 
    \item Feed-forward torque loss weighting: 0.25
\end{itemize}

\subsection{Velocity matching metric}
The velocity metric $V$ is calculated based on the desired x, y, and yaw changes over $n_{ex}$ time steps, using the weights from Sec. \ref{sec:policy_appendix} as follows,
\begin{equation}
\begin{aligned}
   e_1 ={}& w_x(\Delta x_{des} - \Delta x)^2 + \\
     &w_y(\Delta y_{des} - \Delta y)^2 + \\
     &w_\gamma(\Delta \gamma_{des} - \Delta \gamma)^2 \\
     e_2 ={}& w_x(\Delta x_{des})^2 +
    w_y(\Delta y_{des})^2 +
    w_\gamma(\Delta \gamma_{des})^2 \\
     V ={}& (e_2 - e_1)/e_2.
\end{aligned}
\end{equation}

\subsection{MLP baseline}

\subsubsection{Hyperparameters}
The MLP baselines were set with a comparable depth and capacity as the GNNs.
The same hidden layer sizes were used for both the ``shared trunk'' and ``hardware-conditioned'' baselines.
The shared trunk network has a separate input and output layer for each design, which transforms its inputs and outputs to the hidden layer dimension. 
The hardware-conditioned network uses the same input and output layers for all designs, but pads the inputs with zeros as needed to reach the input layer size.
\begin{itemize}
    \item Model network, 6 hidden layers with 300 ReLU units
    \item Policy network, 6 hidden layers with 350 ReLU units
\end{itemize}

\subsubsection{Zero-shot transfer comparison}
\begin{figure}[tb] 
     \centering
      \includegraphics[width=.95\linewidth]{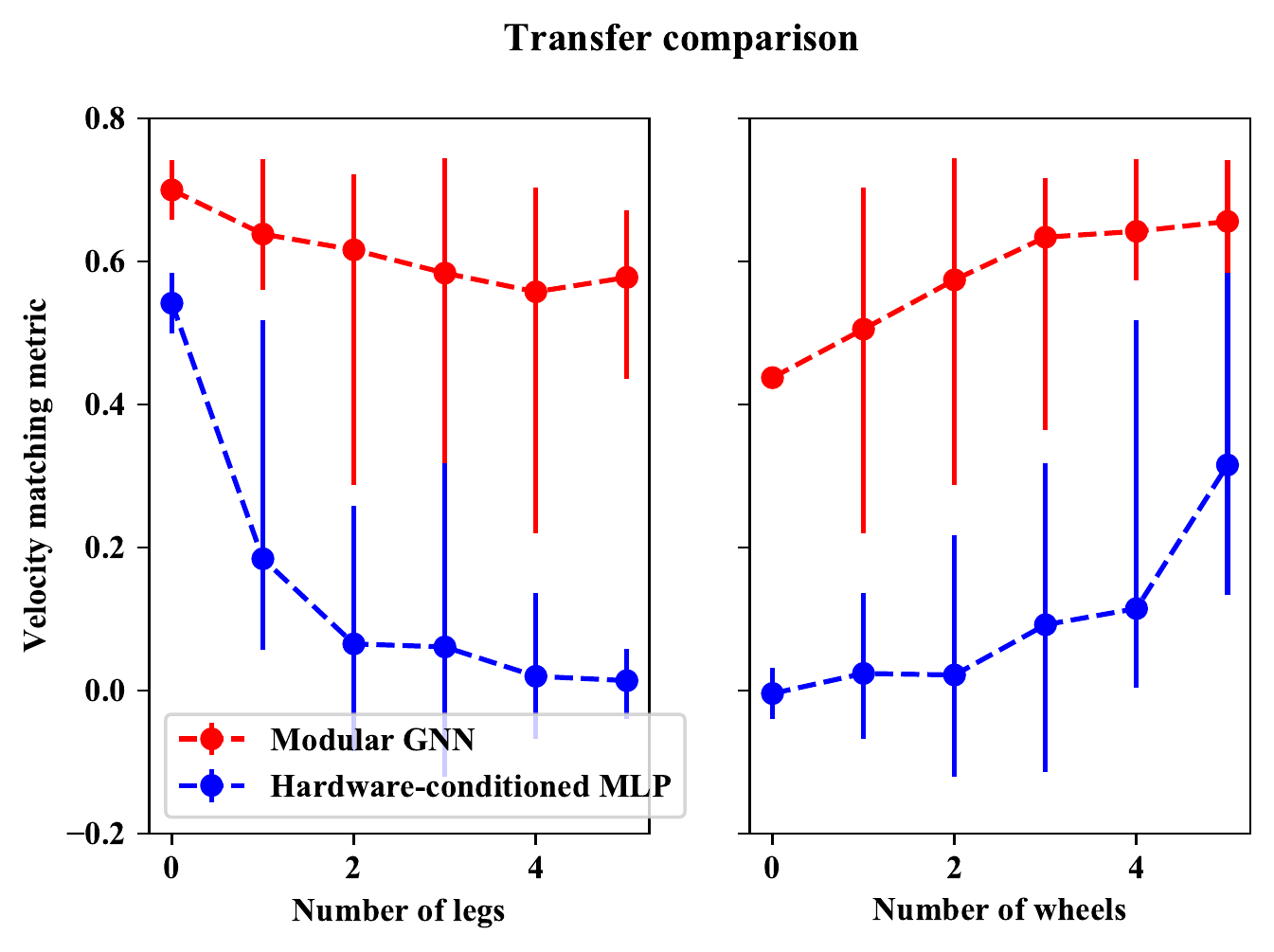}
      \caption{The zero-shot transfer test results from Sec. \ref{sec:mlp_results} and Fig. \ref{fig:transfer_boxplot}, broken down by number of leg and wheel modules in each robot design.
      Multiple designs may have the same number of legs and/or wheels, arranged in different ways.
      The circle markers show the mean of the designs with a given number of legs or wheels, and the vertical bars indicate the max and min. 
      The velocity matching metric measures how well the robot can match a desired heading and speed using the learned policy.
      As the number of legs increases and number of wheels decreases, the MLP policy performance degrades significantly more than does the GNN policy.
      In all cases, our GNN policy is able to transfer more effectively to new designs.  
      }
      \label{fig:transfer_compare}
\end{figure}

In Sec. \ref{sec:mlp_results} with corresponding Fig. \ref{fig:transfer_boxplot}, we found that our GNN was able to generalize to unseen designs more effectively than a hardware-conditioned MLP.
However, the average, max and min performance of the many designs does not reveal which designs the policy can transfer to. We plot the number of leg modules and wheel modules against the transfer results in Fig. \ref{fig:transfer_compare}.
We can see that the MLP is able to transfer most effectively to designs with wheels, which agrees with our intuition that wheels are ``easier'' to learn to control.

\bibliography{refs}

\end{document}